\documentclass{article}

\PassOptionsToPackage{numbers, compress}{natbib}

\usepackage[preprint]{neurips_2026}

\usepackage[utf8]{inputenc} 
\usepackage[T1]{fontenc}    
\usepackage{hyperref}       
\usepackage{url}            
\usepackage{booktabs}       
\usepackage{amsfonts}       
\usepackage{nicefrac}       
\usepackage{microtype}      
\usepackage{xcolor}         
\usepackage{xspace}

\usepackage{algorithm}
\usepackage{algpseudocode}
\usepackage{todonotes}
\usepackage{graphicx}
\usepackage{amsmath}
\usepackage{multirow}
\usepackage{booktabs}
\usepackage{cleveref}
\usepackage{wrapfig}

\crefname{section}{Sec.}{Secs.}
\crefname{figure}{Fig.}{Figs.}

\newcommand{\ours}{Beaver\xspace}

\newcommand{\std}[1]{{\scriptsize\textcolor{gray}{($\pm$#1)}}}

\newlength\savewidth
\newcommand\shline{%
  \noalign{\global\savewidth\arrayrulewidth
  \global\arrayrulewidth 1pt}%
  \hline
  \noalign{\global\arrayrulewidth\savewidth}%
}

\title{Building Agent Harnesses for\\ Scientific Curation from Multimodal Sources}

\author{%
  Sheng Zhang$^{1}$\thanks{Equal contribution.}\hspace{0.6em}
  Qin Liu$^{1,2}$\footnotemark[1] \vspace{0.1em}\\ 
  \textbf{Renqian Luo}$^{1}$\hspace{0.6em}
  \textbf{Shufang Xie}$^{1}$\hspace{0.6em}
  \textbf{Reuben Tan}$^{1}$\hspace{0.6em}
  \textbf{Sean Hayes}$^{3}$\hspace{0.6em}
  \textbf{Gregory Bryman}$^{3}$\\
  \textbf{Wendong Ge}$^{3}$ \hspace{0.6em}
  \textbf{Ruilian Zhang}$^{3}$ \hspace{0.6em}
  \textbf{Oluwaseun Egbelowo}$^{3}$\hspace{0.6em}
  \textbf{Kelly Yee}$^{3}$\hspace{0.6em}
  \textbf{Hoifung Poon}$^{1}$ \vspace{0.2em}\\
  $^{1}$Microsoft Research \quad
  $^{2}$University of California, Davis \\
  $^{3}$Merck \& Co., Inc., Rahway, NJ, USA
}

\begin{document}

\maketitle

\begin{abstract}
Scientific discovery workflows often depend on structured curation from the literature.
This is difficult for current agents because the key evidence is scattered across long text, dense tables, and figures, and the final records often require reasoning across multiple evidence fragments rather than copying a single span.
We study scientific curation from multimodal sources and introduce \ours, an agent harness that extracts structured information from scientific papers while preserving provenance to the supporting evidence.
\ours combines a frontier agent with multimodal evidence tooling, task scaffolding, and artifact-grounded autoresearch.
These components turn curation into a staged, auditable workflow and enable an iterative evaluate--diagnose--revise loop, where persistent run artifacts expose stage-localized failures and guide harness updates.
Experiments show that \ours reaches 81.0 on Gold-Referenced Attribute Score (GRAS), an attribute-level measure of agreement with gold curated records, outperforming frontier agents by over 23 absolute points.
Ablations show that task scaffolding, multimodal evidence tooling, and provenance traces each contribute meaningfully to performance, while attribute-level analysis shows the largest gains on high-value attributes that require cross-modal reasoning and normalization.
These results show that, for scientific curation from papers with multimodal evidence, harness design is a central determinant of agent performance.
\end{abstract}
\section{Introduction}
\label{sec:introduction}

\begin{wrapfigure}{r}{0.35\linewidth}
    \centering
    \vspace{-1em}
    \includegraphics[width=\linewidth]{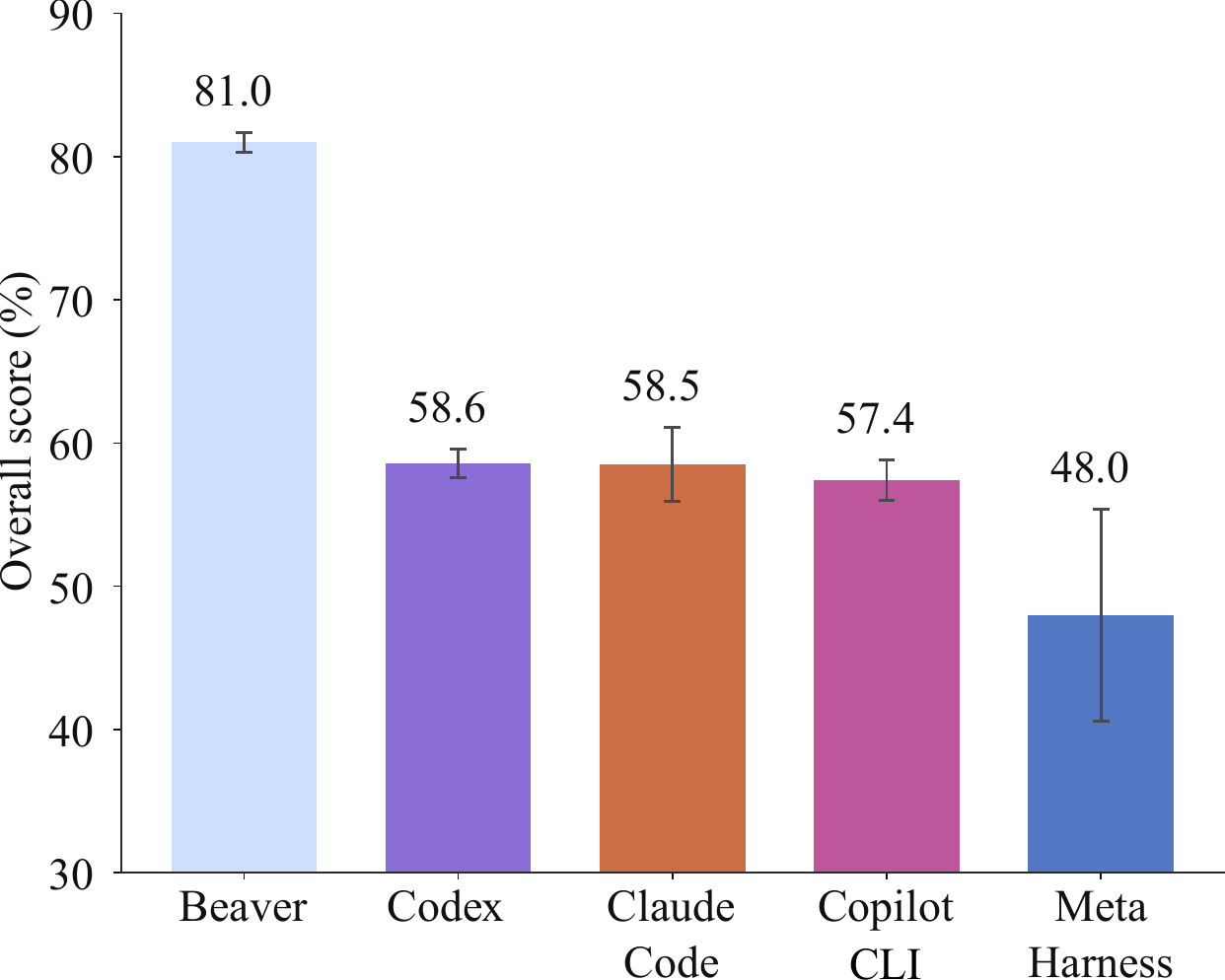}
    \vspace{-1.5em}
    \caption{Performance comparison on multimodal scientific curation. }
    \label{fig:fast_subset_overall_baselines}
\end{wrapfigure}

Scientific discovery workflows often depend on structured curation from the literature: converting papers into normalized records that can be searched, compared, and reused in downstream analyses~\cite{bai-etal-2024-schema,padmakumar-etal-2025-intent}. This need is especially acute in drug discovery and other evidence-intensive domains, where literature curation is often the first step toward building modeling datasets and informing later decisions~\cite{chan2022applications,asai2026synthesizing}. The same pattern arises more broadly whenever experts must integrate prose, tables, and figures into a common set of schema attributes~\cite{padmakumar-etal-2025-intent,ghosh-etal-2024-toward,matvix}. Yet manual curation remains difficult to scale~\cite{lu2012biocuration,arighi2011biocreative}. Real-world scientific curation still relies on domain experts to extract relationships from the literature into systematically structured records~\cite{ctd_pubtator_2025}, creating an operational bottleneck in scientific pipelines with economic stakes on the order of hundreds of millions of dollars annually~\cite{certara_q42025}.

Therefore, it is essential to characterize the task accurately. Scientific structured curation is not a standard document question-answering problem~\cite{lee2023qasa,dasigi-etal-2021-dataset}. The required evidence is scattered across long text, dense tables, and figures, and the final records often require reasoning across multiple evidence fragments rather than copying a single span. For example, constructing one trial record may require linking an intervention and arm described in the methods to outcomes reported in tables or plots, then normalizing the endpoint, unit, value, and sample count into the target schema. Prior systems \cite{scidasynth,axcell,matvix} reflect growing interest in structured extraction from scientific papers, but they also underscore that the core difficulty is not short-form answer generation alone. The task requires multimodal evidence access, cross-source reasoning, schema-aware normalization, and provenance traces that make each extracted value auditable.


Recent frontier agents can accelerate early-stage curation and triage~\cite{asai2026synthesizing,wang2025foundation}, but reliable stand-alone scientific curation remains unresolved. In practical settings, errors are costly because extracted records are consumed by downstream analyses rather than merely read as natural-language summaries. Our analysis (\cref{sec:results,sec:analysis}) reveals a consistent failure pattern: frontier agents often recover high-level study metadata more reliably than high-value quantitative attributes, while tables and figures remain a major source of brittleness. These failures suggest that scientific curation is a long-horizon agent task requiring document navigation, evidence alignment, normalization, and auditability.

Existing public evaluations do not fully capture this setting. Many scientific-document benchmarks emphasize table detection, table-structure recovery, claim verification, or lightweight result extraction, rather than full-paper curation into a rich production-style schema~\cite{zhao-etal-2025-multimodal-foundation,pubtables-1m,wadden-etal-2020-fact}. Meanwhile, recent work on agent harnesses shows that agent performance depends substantially on the control logic and runtime environment surrounding the model, not only on the model itself~\cite{meta-harness,nl-agent-harnesses,harness-native-software-engineering,anthropic-harness-design}. This observation is particularly relevant for scientific curation: when success depends on iterative document navigation, multimodal evidence handling, provenance tracking, and structured output constraints, the surrounding harness can determine whether a strong base agent succeeds or fails.

To study this problem, we introduce a benchmark for scientific curation from papers with multimodal evidence and present \textbf{\ours}, an agent harness for extracting structured information while preserving provenance to the supporting evidence. \ours combines three interacting components: \emph{multimodal evidence tooling}, \emph{task scaffolding}, and \emph{artifact-grounded autoresearch}. Rather than treating the base model as the only locus of improvement, \ours makes the harness itself the object of iterative improvement. It decomposes curation into controllable stages, provides interfaces for textual and visual evidence, and runs an evaluate--diagnose--revise loop in which persistent run artifacts expose stage-localized failures and guide harness updates. Experiments show \ours substantially outperforms frontier agents (\Cref{fig:fast_subset_overall_baselines}), demonstrating that performance on scientific curation depends not only on the base agent, but also on how the harness is organized and improved from execution feedback.
\section{Task}
\label{sec:task}

\begin{figure*}[!ht]
\centering
\includegraphics[width=0.92\textwidth]{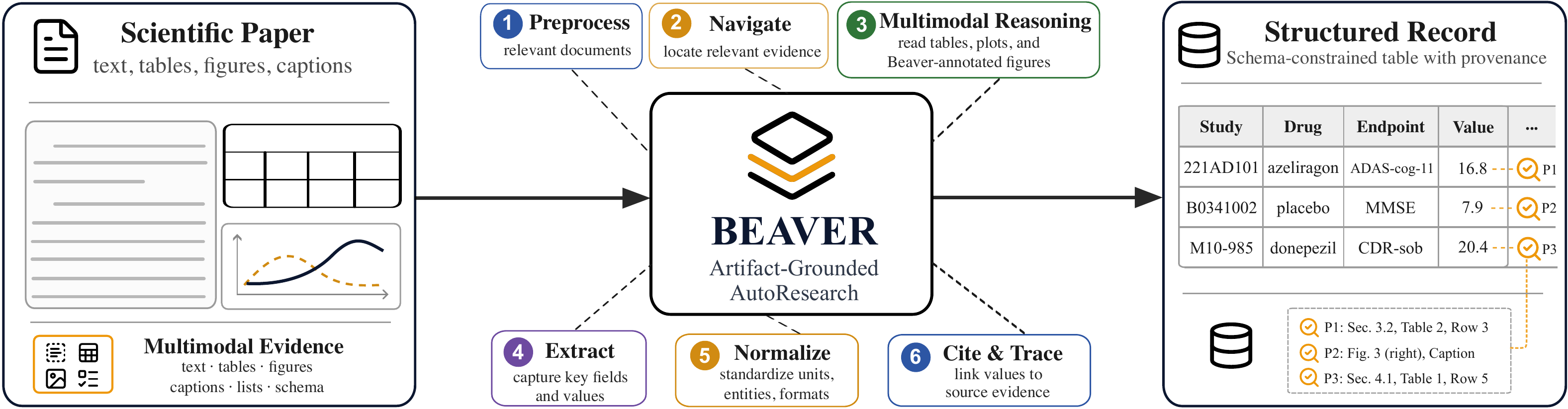}
\caption{Task overview. Given a full scientific paper containing text, tables, and figures, the task is to produce structured records with provenances that trace each value back to supporting source evidence.}
\label{fig:task}
\end{figure*}

\paragraph{Task Definition}

As shown in \cref{fig:task}, we study schema-conditioned extraction from full scientific papers. For a paper $p$, let $\mathcal{T}(p)$ denote its textual content, $\mathcal{B}(p)$ its tables, and $\mathcal{G}(p)$ its figures. Let $\mathcal{F} = \{f_1, \ldots, f_m\}$ be the canonical attribute set defined by the curation schema, including attribute names, normalization requirements, value descriptions, etc. The input to the task is therefore not just a document in the abstract, but the tuple $(\mathcal{T}(p), \mathcal{B}(p), \mathcal{G}(p), \mathcal{F})$.
The output is a predicted table $\hat{Y}$ whose columns are drawn from $\mathcal{F}$ and whose rows correspond to structured records supported by the paper. The task therefore requires the system to populate a schema-constrained table rather than emit free-form text or a question-answer pair, and it assumes that extracted values must follow the canonical attribute definitions and normalization conventions associated with $\mathcal{F}$.
In addition to the predicted table, the task also outputs $\hat{C}$, a provenance artifact that links extracted content in $\hat{Y}$ back to supporting evidence in the paper. While these provenance traces are not used in evaluation,  they make the extracted table auditable and provide grounded evidence for failure analysis during harness development; \Cref{tab:ablation-results} in \Cref{sec:results} reports an ablation study.

\begin{table*}[t]
\centering
\small
\setlength{\tabcolsep}{2.5pt}
\renewcommand{\arraystretch}{1.4}
\newcommand{\cmark}{\textcolor{green!60!black}{\ensuremath{\checkmark}}}
\newcommand{\xmark}{\textcolor{red!70!black}{\ensuremath{\times}}}
\newcommand{\scorecell}[3]{\makebox[4.2em][l]{\rlap{\raisebox{0.15ex}{\scriptsize #1}}\textcolor{#3}{\raisebox{-0.45ex}{\rule{#2}{0.7ex}}}}}
\begin{tabular}{l c c c c c | c c}
\shline
\multirow{2}{*}{\textbf{Benchmark}} & \multirow{2}{*}{\textbf{Full Paper}} & \multirow{2}{*}{\textbf{Multimodal}} & \textbf{Rich} & \textbf{Expert} & \textbf{Long-Horizon} & \textbf{Codex} & \textbf{Claude Code} \\
 & & & \textbf{Schema} & \textbf{Gold} & \textbf{Curation} & {\scriptsize (GPT-5.4)} & {\scriptsize (Opus 4.7)} \\
\hline
PubTables-1M~\cite{pubtables-1m} & \xmark & \xmark & \xmark & \xmark & \xmark & 87.50 & 96.50 \\
SciTSR~\cite{scitsr} & \xmark & \xmark & \xmark & \xmark & \xmark & 97.10 & 96.01 \\
SciER~\cite{zhang-etal-2024-scier} & \cmark & \xmark & \xmark & \xmark & \xmark & 80.45 & 84.62 \\ 
SciClaimEval~\cite{sciclaimeval} & \cmark & \cmark & \xmark & \xmark & \xmark & 92.38 & 100.00 \\
\hline
\textbf{Our benchmark} & \cmark & \cmark & \cmark & \cmark & \cmark & 58.60 & 58.50 \\
\shline
\end{tabular}
\caption{Benchmark-design comparison with related datasets. ``Full Paper'' means the input is the entire paper rather than a page crop or isolated artifact. ``Multimodal'' means the benchmark explicitly requires evidence from text, tables, and figures. ``Rich Schema'' means the target is a long, normalized, production-style field inventory intended for downstream scientific curation rather than a lightweight structured output. ``Expert Gold'' means the benchmark is backed by professionally curated reference data produced on a real curation platform rather than generic labels or weak supervision. ``Long-Horizon Curation'' means the task requires multi-step scientific curation rather than one-shot extraction or classification. The baseline score columns report representative frontier-model performance on each benchmark.}
\label{tab:benchmark_comparison}
\end{table*}

\paragraph{Benchmark Construction}

Existing benchmarks do not match the setting we study here: full-paper scientific curation with evidence distributed across text, tables, and figures, a rich production-style schema, expert-curated reference data, and long-horizon extraction requirements. As summarized in ~\Cref{tab:benchmark_comparison}, related benchmarks miss one or more of these requirements. At the same time, frontier agents already saturate these benchmarks, which further limits their usefulness as tests of real-world scientific curation.

To fill this gap, we construct our benchmark from a professionally curated corpus of PubMed articles. We hired expert curators
to annotate the papers into a structured tabular resource. The resulting corpus focuses on Alzheimer's disease, which spans publications from 1990 to 2023 and contains 425 references, 327 studies, 929 study arms, 106{,}257 patients, and 58{,}058 curated data rows. Within this disease area, it covers
143 interventions, 69 efficacy endpoints, and 373 safety endpoints, providing substantial clinical and structural breadth.

Within this benchmark, the curation schema has over 400 attributes, which are organized at a high level around identifiers, treatment descriptors, covariates, endpoint metadata, and reported outcome values. This organization reflects the practical structure of downstream scientific curation workflows: the benchmark is not asking for a small set of extracted facts, but for a normalized record that combines study context, cohort context, intervention details, and measured results. We select a list of high-value attributes for experiment in \Cref{sec:setup} and provide their definitions in Appendix~\ref{app:schema-details}.


\paragraph{Evaluation Metrics}

Our goal is to evaluate whether a system constructs the correct normalized curation records, not whether it reconstructs the layout of a source table. Existing table metrics~\cite{teds,grits,scitsr} are mainly for source-table detection or structure recovery. Our setting instead compares a predicted curation table against a professionally curated reference table, so the evaluation must be attribute-aware and gold-referenced rather than layout-aware; in that sense it is closer to the recent ExtractBench~\cite{extractbench}, while remaining specialized to scientific curation from multimodal sources. Concretely, we first align predicted and gold rows within each paper using maximum-weight bipartite matching under an attribute-aware row-similarity score. We then compute attribute scores with type-specific functions and report a single gold-referenced aggregate that averages those scores over all gold rows, so unmatched gold rows are penalized through one-sided missingness. Below we formalize this evaluation procedure.

For a paper $p$, let $\hat{R}(p) = \{\hat{r}_1, \ldots, \hat{r}_{n_p}\}$ denote the predicted rows and $R^\star(p) = \{r^\star_1, \ldots, r^\star_{m_p}\}$ the gold rows, where each row is indexed by the evaluated attribute subset $\mathcal{F}_{\mathrm{eval}} \subseteq \mathcal{F}$. We align rows only within the same paper. Given an attribute score $s_f(\hat{r}, r^\star)$ for each $f \in \mathcal{F}_{\mathrm{eval}}$, row similarity is
\[
\sigma(\hat{r}, r^\star) = \frac{1}{|\mathcal{F}_{\mathrm{eval}}|} \sum_{f \in \mathcal{F}_{\mathrm{eval}}} s_f(\hat{r}, r^\star).
\]
We then choose a maximum-weight bipartite matching
\[
A_p = \arg\max_{A} \sum_{(\hat{r}, r^\star) \in A} \sigma(\hat{r}, r^\star)
\]
using Hungarian assignment. Unmatched gold rows are retained in evaluation and paired with null predictions, while prediction-only rows are excluded from the headline aggregate.

Attribute scores are defined by attribute type. For exact-match attributes, we use $s_f(\hat{r}, r^\star) = \mathbf{1}[\hat{r}_f = r^\star_f]$.
For lexical attributes such as drug names, we use token-level F1. 
For numeric attributes such as durations, and outcome values, we use relative accuracy
$s_f(\hat{r}, r^\star) = \max(0, 1 - |\hat{r}_f - r^\star_f|/|r^\star_f|)$

%

For each aligned gold row $r^\star$, we define its row score as the mean attribute score
\[
S(\pi(r^\star), r^\star) = \frac{1}{|\mathcal{F}_{\mathrm{eval}}|} \sum_{f \in \mathcal{F}_{\mathrm{eval}}} s_f(\pi(r^\star), r^\star),
\]
where $\pi(r^\star)$ is the matched predicted row, or a null row if $r^\star$ is unmatched. Our headline metric is the \emph{Gold-Referenced Attribute Score},
\[
\mathrm{GRAS} = \frac{1}{\sum_p |R^\star(p)|} \sum_p \sum_{r^\star \in R^\star(p)} S(\pi(r^\star), r^\star).
\]
Because unmatched gold rows remain in the denominator and incur one-sided missingness penalties, this aggregate jointly reflects row recovery and attribute correctness while keeping the evaluation anchored to the professionally curated reference rows.

\section{Method}
\label{sec:method}

\begin{figure*}[!ht]
\centering
\includegraphics[width=0.98\textwidth]{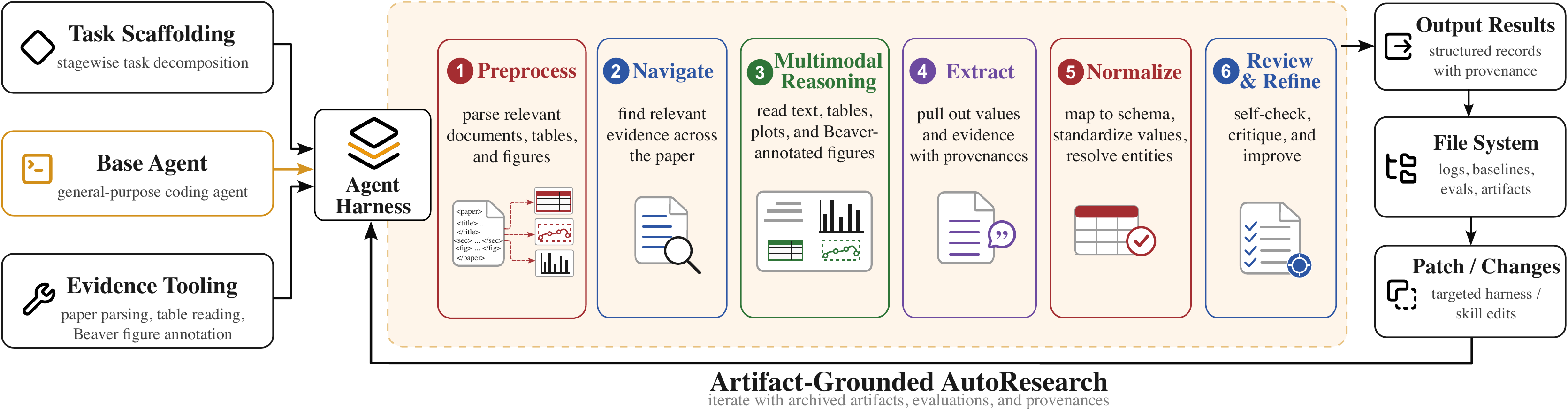}
\caption{Method overview. Task scaffolding, a base agent, and multimodal evidence tooling form an initial agent harness. During artifact-grounded autoresearch, the harness executes staged curation workflows, writes outputs, logs, evaluations, and provenance traces to the filesystem, and uses these artifacts to propose, test, and retain targeted harness changes.}
\label{fig:method_panels}
\end{figure*}


\begin{algorithm}[t]
\caption{Artifact-grounded autoresearch for harness optimization}
\label{alg:autoresearch}
\begin{algorithmic}[1]
\Require Initial harness $H_0=(\{\omega_k\}_{k=1}^K,\mathcal{U},\Pi)$, dev set $\mathcal{D}_{\mathrm{dev}}$, stages $1{:}K$
\For{$t=0,1,\ldots,T-1$}
    \State $R_t \gets \Call{EvaluateByStage}{H_t,\mathcal{D}_{\mathrm{dev}}}$
    \State $k^\star \gets \arg\min_k\; \Call{Score}{R_t,k}$
    \State $\mathcal{S}_{t,k^\star} \gets \Call{InitWorkspace}{R_t,k^\star}$ \Comment{tables, logs, summaries, trajectories, etc.}
    \For{$j=1,\ldots,J$}
        \State $\Delta_{t,j} \gets \Call{ProposePatch}{H_t,\mathcal{S}_{t,k^\star}}$
        \State $H' \gets \Call{ApplyPatch}{H_t,\Delta_{t,j}}$ \Comment{update the harness: scaffold, tools, scripts, etc.}
        \State $O_{t,j} \gets \Call{RunStage}{H',k^\star,\mathcal{D}_{\mathrm{dev}}}$ \Comment{papers follow the updated workflow}
        \State $\mathcal{S}_{t,k^\star} \gets \Call{UpdateWorkspace}{\mathcal{S}_{t,k^\star},O_{t,j}}$
        \State $H_t \gets \Call{AcceptOrDiscard}{H_t,H',\mathcal{S}_{t,k^\star}}$ \Comment{keep only if better}
    \EndFor
    \State $H_{t+1} \gets H_t$
\EndFor
\State \Return $H_T$
\end{algorithmic}
\end{algorithm}

\ours is a harness for scientific curation from multimodal sources built around a base agent and three interacting components: multimodal evidence tooling, task scaffolding, and artifact-grounded autoresearch. Our method centers on how the harness exposes scientific evidence, structures the task, and iteratively revises itself using grounded diagnosis. \Cref{fig:method_panels} illustrates this framework.

For a paper $p$ and the canonical extraction schema introduced in \cref{sec:task}, \ours runs a staged extraction process. Stages are indexed by $k \in \{1, \ldots, K\}$, and each stage predicts a cumulative table $\hat{Y}^{(k)}$ that extends the attributes completed by earlier stages. The harness exposes multimodal evidence tools $\mathcal{U}$, which produce and operate on derived artifacts such as markdown sections, parsed tables, and calibrated figure-reading artifacts. These tools are used through a stage-specific workflow scaffold $\omega_k$, which guides document preprocessing, evidence navigation, multimodal reasoning, extraction, etc. Over autoresearch iterations $t = 1, 2, \ldots$, the harness $H_t = (\{\omega_k\}_{k=1}^K, \mathcal{U}, \Pi)$ is revised using archived run artifacts and evaluation feedback, where $\Pi$ denotes the prompts, scripts, validators, and other implementation components of the harness.

\textbf{Base Agent.} At the base of \ours is a frontier agent that provides state-of-the-art general agentic capacity. This base agent is already capable of acting as an autonomous agent that reads, modifies, and executes code directly in a local environment. However, the base agent alone is not sufficient on this benchmark and performs substantially worse than the full \ours harness. Our contribution is therefore a harness that makes the same underlying agent more effective on scientific curation from multimodal sources. The detailed performance comparison is reported in \Cref{tab:main-results} in \Cref{sec:results}.

\textbf{Multimodal Evidence Tooling.} To improve multimodal evidence access, \ours is equipped with an auxiliary tool set $\mathcal{U}$. This includes a paper-to-markdown transformation that turns each article into an outline plus section-level markdown files, which makes long documents easier to navigate than raw XML or PDF alone. It also includes a parsed table viewer and a calibrated gridline overlaying tool that help the agent recover quantitative values from plots. During scaffolded stage execution, these tools are invoked as the agent moves through the workflow (\Cref{fig:method_panels}): preprocessing prepares usable evidence artifacts, navigation selects where to look, multimodal reasoning reads the relevant text, table, or figure artifact, and extraction records values together with provenance traces. The tools do not solve extraction directly, but they make the scaffolded actions described next possible and more accurate by providing a better interface to the textual, tabular, and visual evidence.

\textbf{Task Scaffolding.} Rather than asking the agent to solve the full curation problem in one pass, \ours decomposes extraction into sequential stages that move from easier paper-level attributes to harder arm-level and endpoint-level attributes. Each stage is not only an attribute subset, but also a workflow scaffold. As shown in \Cref{fig:method_panels}, a typical stage asks the agent to preprocess relevant documents, tables, and figures; navigate to candidate evidence; perform multimodal reasoning and annotation over text, tables, and plots; extract values with provenance traces; normalize entities and units into the schema; and review and refine the output. These scaffolded actions play a critical role in the harness: they reduce unproductive freedom, make the agent spend computation on evidence-bearing operations, and improve extraction quality (See \Cref{tab:ablation-results} in \Cref{sec:results} for the details).
Formally, stage $k$ predicts
\[
\hat{Y}^{(k)} = h_k\!\left(p, \mathcal{A}(p), \hat{Y}^{(k-1)}; \omega_k, \mathcal{U}\right),
\]
where $\hat{Y}^{(0)}$ is empty, $\omega_k$ specifies the stage workflow, and $\mathcal{U}$ exposes the evidence tools available to the agent during that workflow. Later stages consume the carried-forward outputs of earlier stages, which reduces ambiguity and narrows the search space for each step. This staged formulation improves initial tractability, but it is equally important for iteration: it lets the harness diagnose and revise failures at a specific stage without conflating unrelated errors elsewhere in the pipeline.

\textbf{Artifact-Grounded Autoresearch.} The base agent, multimodal evidence tools, and task scaffolding together form the initial agent harness $H_0$. Artifact-grounded autoresearch then improves this harness through persistent filesystem state. Unlike a free-form self-improvement loop~\cite{karpathy-autoresearch,meta-harness}, \ours uses the stage structure from task scaffolding to localize failures. \Cref{alg:autoresearch} summarizes our two-level optimization procedure. At the start of each outer iteration, the current harness is evaluated on a development set $\mathcal{D}_{\mathrm{dev}}$. The evaluation is decomposed by stage using the task scaffold, the lowest-performing stage is selected, and a stage-specific workspace is initialized. This workspace contains a living filesystem, including progress tables, logs, run summaries, raw outputs, agent execution trajectories, provenance traces, and other artifacts needed for grounded comparison.

The inner loop focuses improvement on the selected stage. At each step, it first resets the working harness to the current best checkpoint. Next, it reviews the stage workspace, proposes targeted changes, and applies a candidate patch to the harness. The candidate harness then reruns the focused stage on $\mathcal{D}_{\mathrm{dev}}$. Because the patch may change $\omega_k$, $\mathcal{U}$, or $\Pi$, the same papers may now pass through an updated sequence of actions, such as different preprocessing, navigation, figure reading, normalization, or review procedures. Once outputs are produced, the filesystem is updated with new results, logs, trajectories, and summaries, and the loop repeats. Based on these artifacts and the resulting stage performance, the revision is either accepted as the new current best harness or discarded. The next inner-loop step therefore proposes changes from the best accepted harness while still retaining evidence from unsuccessful attempts. This artifact-grounded state lets the loop build on prior evidence without accumulating unsuccessful patches.

Overall, the base agent supplies general-purpose agentic capacity; multimodal evidence tooling gives the agent reliable interfaces to text, tables, and figures; task scaffolding organizes those capabilities into controllable stage-level workflows; and artifact-grounded autoresearch uses the resulting artifacts to identify, test, and retain harness improvements. Compared with free-form generic harness-improvement methods~\cite{meta-harness}, this stage-localized, artifact-grounded loop is better matched to scientific curation from multimodal sources, where failures often arise from specific evidence-processing steps rather than from the end-to-end task alone. As shown in \Cref{sec:results}, this design yields substantially stronger performance than both related methods and ablated variants.
\section{Experiment Setup}
\label{sec:setup}

\paragraph{Evaluation Corpus and Schema}

We evaluate the benchmark introduced in \cref{sec:task} on a selected set of 23 papers drawn from PMC Open Access~\cite{pmc_open_access_subset}.
We restrict evaluation to this set because full-paper access is available for every article, which allows each compared harness to operate over the complete paper rather than over abstracts or partial artifacts. Appendix~\ref{app:evaluation-papers} lists the PMIDs in the evaluation. Instead of the full attribute schema, we select 20 high-value attributes for multimodal extraction evaluation. These attributes capture the most important extraction targets for drug-discovery and clinical-trial analysis while keeping evaluation tractable and aligned across all compared systems. Appendix~\ref{app:schema-details} provides the detailed definitions of these attributes.


\paragraph{Development Set}

For harness development, we use a smaller subset of 8 papers sampled from the full evaluation set. This subset makes iteration faster than full-set evaluation while still covering representative failure modes and stable anchor cases. Appendix~\ref{app:evaluation-papers} marks the development subset.

\paragraph{Baselines}

We compare against four harness baselines: Codex CLI, Claude Code, Copilot CLI, and Meta-Harness~\cite{meta-harness}. The three CLI systems are frontier off-the-shelf agent harnesses that are already used in practical agent workflows, while Meta-Harness is the state-of-the-art open-source harness on Terminal Bench 2.0~\cite{terminalbench}. Across these harnesses, we use frontier LLMs including GPT-5.4, Claude Opus 4.7, Opus 4.6, and Sonnet 4.6. This setup
makes the comparison focus on harness design under realistic high-capability model conditions rather than on differences caused by weak base models.

Our harness uses Copilot CLI as the base agent and GPT-5.4 as the backend model. The task scaffolding decomposes each task into 4 stages. For artifact-grounded autoresearch, we set the number of outer iterations to $T=3$ and the number of inner-loop iterations to $J=10$. This results in 34 iterations in total. Each harness is run three times, and we report the mean and standard deviation across runs. Repeated runs are important because agent executions exhibit meaningful stochastic variance, making single-run comparisons unreliable for evaluating harness-level improvements.
We use Gold-Referenced Attribute Score (GRAS) as the primary metric. In addition, we report attribute-wise averages and stage-wise scores to characterize where each harness succeeds or fails.

\section{Results}
\label{sec:results}

\begin{wraptable}{r}{0.40\textwidth}
\vspace{-1.5em}
\centering
\small
\caption{Main comparison on the evaluation benchmark. Scores are GRAS; gray parentheses show standard deviation.}
\label{tab:main-results}
\vspace{0.5em}
\begin{tabular}{l l c}
\toprule
Harness & Model & Score \\
\midrule
\textbf{\ours} & gpt-5.4 & \textbf{81.0} \std{0.7} \\
Codex & gpt-5.4 & 58.6 \std{1.0} \\
Claude Code & opus-4.7 & 58.5 \std{2.6} \\
Claude Code & opus-4.6 & 58.0 \std{1.3} \\
Copilot CLI & gpt-5.4 & 57.4 \std{1.4} \\
Meta-Harness & gpt-5.4 & 48.0 \std{7.4} \\
Claude Code & haiku-4.5 & 44.0 \std{6.2} \\
\bottomrule
\end{tabular}
\end{wraptable}

\paragraph{Main Comparison}
\Cref{tab:main-results} reports the main comparison on the evaluation benchmark. \ours achieves the highest Gold-Referenced Attribute Score (GRAS), reaching 81.0\%, while the strongest baseline reaches only 58.6\%. This corresponds to a 22.4-point absolute improvement over Codex with the same GPT-5.4 backend. The comparison with Copilot CLI is also informative: because \ours uses Copilot CLI as its base agent, the 23.6-point gap between Copilot CLI and \ours isolates the effect of the proposed harness design rather than a change in the underlying agent interface. Claude Code with stronger Opus backends performs similarly to Codex, suggesting that simply swapping to another high-capability frontier agent is not sufficient for this benchmark. Meta-Harness performs substantially worse than \ours despite using the same GPT-5.4 backend, indicating that generic harness optimization does not directly solve scientific curation from multimodal sources.

Overall, these results show that performance is not determined only by backend model strength. Instead, the way the agent is structured around the task matters: multimodal evidence access, staged task scaffolding, provenance-grounded execution, and artifact-grounded autoresearch together make the same class of frontier agents substantially more effective for full-paper scientific curation.

\begin{figure*}[!ht]
\centering
\includegraphics[width=0.95\textwidth]{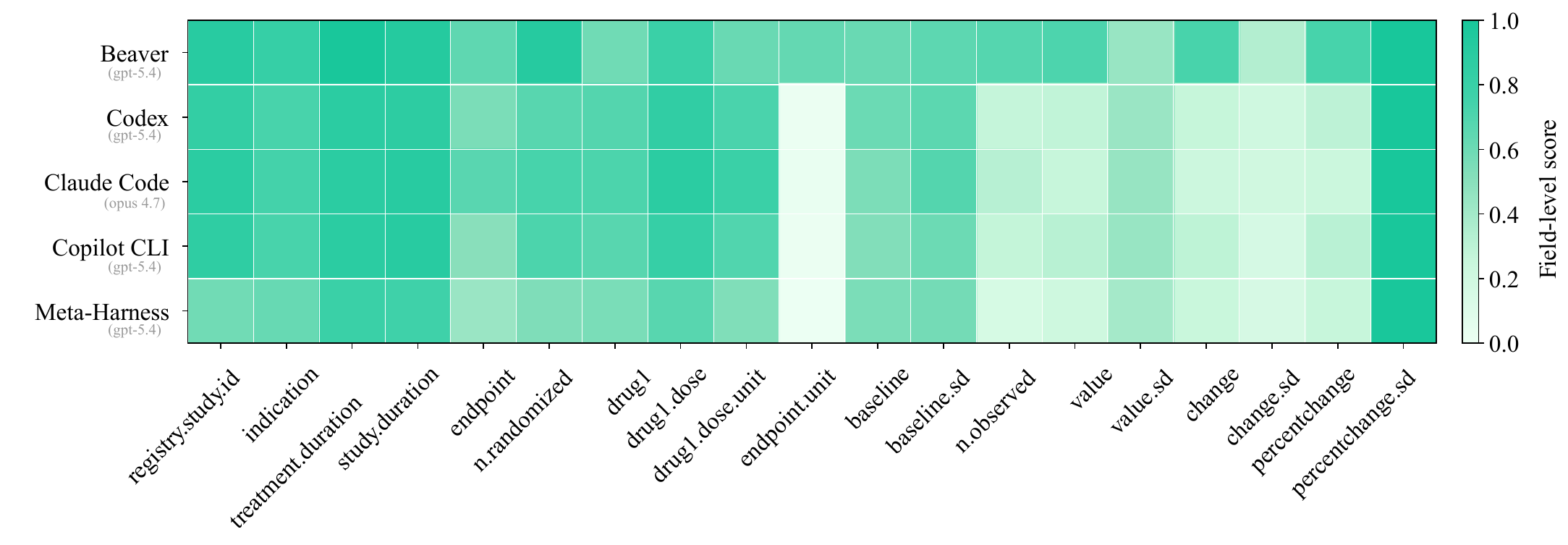}
\vspace{-1.0em}
\caption{Attribute-level score comparison across the main baselines. Rows are harnesses, columns are the displayed scored attributes, and darker green cells indicate stronger attribute-level performance.}
\label{fig:main-field-heatmap}
\end{figure*}

\Cref{fig:main-field-heatmap} breaks the comparison down by attribute. The gains are not confined to a single part of the schema; \ours improves across multiple high-value attributes. The largest advantages appear on attributes such as \texttt{endpoint.unit}, \texttt{change}, \texttt{percentchange}, \texttt{value}, and \texttt{n.observed}. These attributes are difficult for different reasons. \texttt{endpoint.unit} requires schema-aware normalization, while outcome attributes such as \texttt{change}, \texttt{percentchange}, and \texttt{value} often require locating the correct table or figure, aligning arms and endpoints, and interpreting reported measurements rather than copying a nearby span of text. The improvement pattern is therefore consistent with the method design in \cref{sec:method}: staged extraction reduces ambiguity, multimodal tooling improves access to evidence-bearing artifacts, and artifact-grounded autoresearch targets the failure modes that matter most for cross-modal reasoning and normalization.

\paragraph{Ablations}
We ablate the three main components while keeping the rest of the harness fixed. \Cref{tab:ablation-results} shows that all three components contribute substantially to final performance. The full harness reaches 81.0, while removing provenance traces, multimodal evidence tooling, and task scaffolding reduces performance to 70.7, 66.1, and 60.5, respectively. Each of these components changes how effectively the agent can turn multimodal paper evidence into schema-constrained records.

\begin{table*}[!ht]
\centering
\small
\caption{Ablation results for the main components of \ours. GRAS is reported on a 0--100 scale. We report standard deviation for the overall score across repeated runs; stage-wise standard deviations are omitted for readability. Total tokens are measured over a 34-iteration autoresearch budget.}
\label{tab:ablation-results}
\begin{tabular}{lcccccc}
\toprule
System Variant & \multicolumn{5}{c}{Gold-Referenced Attribute Score (GRAS)} & Total Tokens \\ \cmidrule(r){2-6}
 & Overall & Stage 1 & Stage 2 & Stage 3 & Stage 4 & (all iterations) \\ \midrule
\textbf{\ours} & 81.0 \std{0.7} & 91.6 & 79.8 & 73.4 & 79.2 & 743M \\ \midrule
No provenance traces & 70.7 \std{2.1} & 81.1 & 60.9 & 73.0 & 67.9 & 722M \\ 
No multimodal evidence tooling & 66.1 \std{0.8} & 80.1 & 57.3 & 69.1 & 57.9 & 555M \\
No task scaffolding & 60.5 \std{1.4} & - & - & - & - & 700M \\
\bottomrule
\end{tabular}
\end{table*}

The largest drop comes from removing task scaffolding. Without staged workflows, the harness loses both decomposition of the extraction target and the action-level structure that guides the task execution. This variant also has no meaningful stage-wise breakdown in \Cref{tab:ablation-results}, because the stage structure itself has been removed. Its lower final score despite a comparable autoresearch budget indicates that additional agent effort is not enough when the search process is poorly organized.

Removing multimodal evidence tooling causes the next largest degradation. The drop is especially pronounced in the later stages, where extraction depends more heavily on tables, figures, and quantitative evidence. This supports the central motivation for the tooling layer: even a strong base agent benefits from interfaces that expose scientific evidence in forms that are easier to navigate, inspect, and verify. Removing provenance traces produces the smallest but still substantial loss. Since provenance traces are not part of the evaluation metric, this drop suggests that their main value is indirect: provenance traces make runs easier to audit, compare, and debug during autoresearch, which improves the quality of subsequent harness revisions.

\begin{figure*}[!ht]
\centering
\includegraphics[width=0.98\textwidth]{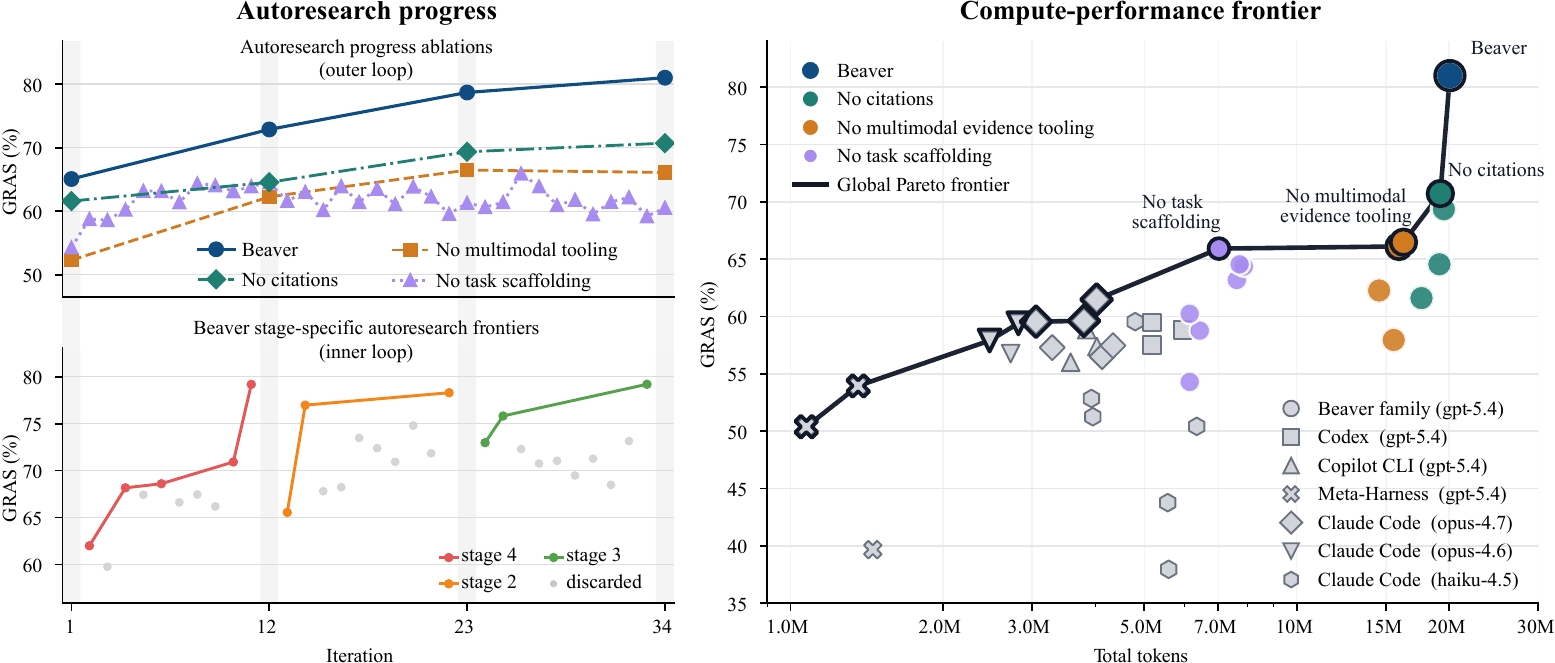}
\caption{
Left: \textbf{Autoresearch progress for \ours and its three ablations.} The upper panel shows all-stage evaluations over autoresearch iterations, including Beaver and three ablations. The lower panel shows \ours’s accepted stage-specific autoresearch frontiers (inner loop), with gray bands indicating the all-stage checkpoint slots (outer loop).
Right: \textbf{Compute--performance frontier.} We plot GRAS against total token usage to compare how efficiently each harness converts agent compute into curation quality. Gray markers denote repeated runs of strong baselines across harnesses and backend models, while colored markers denote \ours and its ablated variants. The connected line shows the empirical Pareto frontier, highlighting the best observed score at each token budget.
}
\label{fig:ablation-progress}
\end{figure*}

\Cref{fig:ablation-progress} (left) complements the final scores by showing how these differences emerge over autoresearch iterations. The full harness improves more reliably over the 34-iteration budget, while each ablation follows a weaker trajectory. The lower panel further shows that stage-localized inner-loop search produces accepted frontiers for the full system, illustrating the mechanism behind the table-level gains: task scaffolding localizes failures, evidence tools make candidate fixes actionable, and provenance-backed artifacts make it easier to decide which patches should be retained.
\ref{fig:ablation-progress-2} in Appendix further visualizes the autoresearch trajectories for each ablated harness.

\paragraph{Compute--Performance Trade-off}

Beyond final accuracy, we ask whether harness improvements convert inference budget into curation quality efficiently. \Cref{fig:ablation-progress} (right) presents a compute--performance frontier: the x-axis measures total token usage and the y-axis measures GRAS. \ours sits at the top of the empirical frontier, achieving the best observed score while using a token budget comparable to other autoresearch variants.
The no-task-scaffolding variant consumes substantial compute but remains far below the full harness, showing that additional agent effort is inefficient when the workflow is poorly organized. Relative to the strongest baseline, this variant approximates adding autoresearch around the base agent without stage-localized task structure. Its gain over the baseline suggests that artifact-grounded search is useful, while its large gap to \ours shows that autoresearch becomes substantially more effective when guided by task scaffolding. The no-provenance-traces and no-multimodal-tooling variants improve with autoresearch, but they plateau below \ours because the search loop has weaker evidence traces or weaker access to multimodal artifacts. 
\section{Analysis}
\label{sec:analysis}

\paragraph{Discovered Harness Changes.}

\ours automatically identifies and refines harness. We find that most improvements fall into three recurring categories.
First, several changes improve \textbf{figure-grounded evidence access}. They replace coarse visual heuristics with more structured reading of figures, enabling consistent extraction of numeric values from plotted trajectories. In our setting, this includes both improved access to section-structured text and explicit mechanisms for mapping visual signals (e.g., plot markers) to calibrated values. A representative example of this figure-reading pipeline is provided in \Cref{app:representative-revisions}. 
Second, a set of changes \textbf{preserve stagewise analysis families and carried-forward context}. These address failures where distinct clinical rows—such as pooled groups, subgroups, or comparison families—are prematurely merged or overwritten. The revised harness maintains these families explicitly across stages, ensuring that downstream extraction operates on the correct endpoint-level context. In practice, this means delaying aggregation and preserving identifiers long enough to recover the appropriate structured records. The core mechanism is formalized in \Cref{alg:stage3-row-family-preservation}.
Third, we observe improvements from \textbf{stronger deterministic postprocessing}. These changes enable reliable recovery of derived outcome attributes when sufficient local evidence is already present in the extracted row. Rather than requiring the model to emit fully normalized outputs, the harness performs lightweight normalization and arithmetic (e.g., computing \texttt{value} or \texttt{percentchange}) to match schema expectations. A concrete example is shown in \Cref{app:representative-revisions}.

\paragraph{Failure Mode Analysis.}

Comparing \ours against strong baseline agents and against ablations surfaces three contrasting case types.
First, baseline agents miss \textbf{multi-visit figure trajectories}: they typically settle for one row per arm at the final visit, while \ours recovers calibrated per-visit rows aligned to gold. On \href{https://pubmed.ncbi.nlm.nih.gov/27756421/}{PMID~27756421}, \ours emits 80 rows matching the gold reference layout while the strongest baseline emits only 15 final-visit rows, producing a \(+0.52\) per-paper GRAS gap (Case A in \Cref{app:representative-failure-cases}).
Second, removing individual \ours components causes targeted regressions: stripping task scaffolding inflates extraction to the wrong endpoint family (\href{https://pubmed.ncbi.nlm.nih.gov/22291741/}{PMID~22291741}, \(-0.54\)), removing provenance grounding collapses subgroup row families (\href{https://pubmed.ncbi.nlm.nih.gov/31884472/}{PMID~31884472}, \(-0.40\)), and removing multimodal tooling reverts plot reading to imprecise visual estimates (PMID~27756421, \(-0.32\)); details in Case B of \Cref{app:representative-failure-cases}.
Third, \ours still collapses \textbf{repeated analysis-population variants} (e.g., FAS, observed-cases, completers) into a single row per (subgroup, endpoint), as on \href{https://pubmed.ncbi.nlm.nih.gov/29154277/}{PMID~29154277} where GRAS remains $0.48$; we treat this direction as the priority growth area (Case C in \Cref{app:representative-failure-cases}).
\section{Related Work}

Existing datasets relevant to our setting fall into several adjacent but incomplete categories. AxCell~\cite{axcell} and SciDaSynth~\cite{scidasynth} are among the closest examples of extracting structured results from scientific papers, but they target lighter-weight result extraction settings than the rich curation schema studied here. MatViX~\cite{matvix} is the closest conceptual multimodal extraction benchmark because it explicitly integrates evidence across text, tables, and figures, but it targets structured JSON extraction in a different scientific domain. CTE~\cite{cte}, PubTables-1M~\cite{pubtables-1m}, SciTSR~\cite{scitsr}, and TabLeX~\cite{tablex} are strong scientific-table benchmarks for contextualized extraction, table detection, structure recovery, or table-centric content extraction, but they do not evaluate end-to-end curation into a normalized record schema.
Scientific information extraction benchmarks such as SciERC~\cite{luan-etal-2018-multi}, SciREX~\cite{jain-etal-2020-scirex}, SciER~\cite{zhang-etal-2024-scier}, SciNLP~\cite{duan-etal-2025-scinlp}, EBM-NLP~\cite{nye-etal-2018-corpus}, and BioRED~\cite{luo2022biored} evaluate entity extraction, relation extraction, contribution extraction, measurement extraction, or biomedical evidence extraction from scientific text, but their targets are primarily entity-relation structures, keyphrases, contribution graphs, or span-level annotations rather than multimodal schema-normalized curation records.
SciClaimEval~\cite{sciclaimeval} and cPAPERS~\cite{cpapers} are also multimodal and paper-grounded, yet their targets are claim verification and question answering rather than structured table construction. Taken together, these benchmarks are highly relevant, but none directly evaluates full-paper multimodal scientific curation into a rich production-style schema with expert-curated reference data.

Related method work is closer in spirit to our harness design than to the benchmark itself. Karpathy's \texttt{autoresearch} project~\cite{karpathy-autoresearch} provides the practical starting point for treating iterative research as a filesystem-backed agent loop, while Meta-Harness~\cite{meta-harness} is the closest direct methodological comparator because it also treats the harness as the object of optimization. Natural-Language Agent Harnesses~\cite{nl-agent-harnesses} and Harness-Native Software Engineering~\cite{harness-native-software-engineering} further support the view that the control logic and runtime environment around the model are first-class methodological choices, and recent engineering accounts likewise argue that harness design materially affects long-horizon agent behavior~\cite{anthropic-harness-design}. More broadly, recent work has explored reflective self-improvement~\cite{self-refine,liu2025metascale}, text-based optimization~\cite{textgrad,feedback-descent}, evolutionary or zeroth-order search~\cite{adaevolve,alphaevolve}, and prompt or pipeline optimization frameworks~\cite{gepa,dspy}. As emphasized in Meta-Harness's related-work framing, many of these methods operate with shorter-horizon or more compressed feedback than full harness development. In contrast, \ours focuses on multimodal scientific curation, combines staged extraction with multimodal evidence artifacts, and grounds revision in archived outputs, provenance traces, evaluation artifacts, diagnosis artifacts, and human-reviewed winning runs.

\section*{Discussion}
\label{sec:discussion}

Our results show that harness design matters in scientific curation from full papers with multimodal evidence. At the same time, the current study leaves several important open directions.

One is scaling the autoresearch loop further, both to test whether performance continues to improve beyond the current 34-iteration horizon and to study when returns begin to flatten. Another is more parallel development. In principle, stage-specific autoresearch can be run concurrently, with cross-stage transferred checkpoints used to test whether local gains compose cleanly at the system level. A third direction is broader generalization. The framework developed here is motivated by scientific curation from multimodal sources, but the underlying pattern of staged task structure, evidence-access tooling, and artifact-grounded harness revision should transfer to other multi-step agent tasks that require structured outputs rather than free-form answers.

More broadly, the main positive impact of this line of work is improved scientific curation efficiency. Many discovery workflows depend on extracting normalized evidence from large volumes of literature, and multimodal papers are a persistent bottleneck because the relevant evidence is distributed across text, tables, and figures. A stronger harness for this setting could reduce manual effort and accelerate downstream analysis. At the same time, this is also a setting where auditability matters: errors in normalized scientific records can propagate into later decisions. For that reason, we view human review, grounded diagnosis, and explicit evidence linkage not as optional extras, but as part of the responsible use of agentic curation systems.

\newpage
\bibliographystyle{unsrtnat}
\bibliography{references}

@article{scidasynth,
author = {Wang, Xingbo and Huey, Samantha L. and Sheng, Rui and Mehta, Saurabh and Wang, Fei},
title = {SciDaSynth: Interactive Structured Data Extraction From Scientific Literature With Large Language Model},
journal = {Campbell Systematic Reviews},
volume = {21},
number = {4},
pages = {e70073},
doi = {https://doi.org/10.1002/cl2.70073},
url = {https://onlinelibrary.wiley.com/doi/abs/10.1002/cl2.70073},
eprint = {https://onlinelibrary.wiley.com/doi/pdf/10.1002/cl2.70073},
year = {2025}
}

@article{ctd_pubtator_2025,
author = {Wiegers, Thomas C. and Davis, Allan Peter and Wiegers, Jolene and Sciaky, Daniela and Barkalow, Fern and Wyatt, Brent and Strong, Melissa and McMorran, Roy and Abrar, Sakib and Mattingly, Carolyn J.},
title = {Integrating {AI}-powered text mining from {PubTator} into the manual curation workflow at the Comparative Toxicogenomics Database},
journal = {Database},
volume = {2025},
pages = {baaf013},
year = {2025},
doi = {10.1093/database/baaf013},
url = {https://doi.org/10.1093/database/baaf013}
}

@misc{matvix,
      title={MatViX: Multimodal Information Extraction from Visually Rich Articles}, 
      author={Ghazal Khalighinejad and Sharon Scott and Ollie Liu and Kelly L. Anderson and Rickard Stureborg and Aman Tyagi and Bhuwan Dhingra},
      year={2024},
      eprint={2410.20494},
      archivePrefix={arXiv},
      primaryClass={cs.CL},
      url={https://arxiv.org/abs/2410.20494}, 
}

@inproceedings{axcell,
    title = "{AxCell}: Automatic Extraction of Results from Machine Learning Papers",
    author = "Kardas, Marcin  and
      Czapla, Piotr  and
      Stenetorp, Pontus  and
      Ruder, Sebastian  and
      Riedel, Sebastian  and
      Taylor, Ross  and
      Stojnic, Robert",
    editor = "Webber, Bonnie  and
      Cohn, Trevor  and
      He, Yulan  and
      Liu, Yang",
    booktitle = "Proceedings of the 2020 Conference on Empirical Methods in Natural Language Processing (EMNLP)",
    month = nov,
    year = "2020",
    address = "Online",
    publisher = "Association for Computational Linguistics",
    url = "https://aclanthology.org/2020.emnlp-main.692/",
    doi = "10.18653/v1/2020.emnlp-main.692",
    pages = "8580--8594"
}

@inproceedings{cte,
  title={CTE: A Dataset for Contextualized Table Extraction},
  author={Gemelli, A and Vivoli, E and Marinai, S and others},
  booktitle={CEUR WORKSHOP PROCEEDINGS},
  volume={3365},
  pages={197--208},
  year={2023},
  organization={CEUR-WS}
}

@misc{pubtables-1m,
      title={PubTables-1M: Towards comprehensive table extraction from unstructured documents}, 
      author={Brandon Smock and Rohith Pesala and Robin Abraham},
      year={2021},
      eprint={2110.00061},
      archivePrefix={arXiv},
      primaryClass={cs.LG},
      url={https://arxiv.org/abs/2110.00061}, 
}

@misc{scitsr,
      title={Complicated Table Structure Recognition}, 
      author={Zewen Chi and Heyan Huang and Heng-Da Xu and Houjin Yu and Wanxuan Yin and Xian-Ling Mao},
      year={2019},
      eprint={1908.04729},
      archivePrefix={arXiv},
      primaryClass={cs.IR},
      url={https://arxiv.org/abs/1908.04729}, 
}

@article{tablex,
  author       = {Harsh Desai and
                  Pratik Kayal and
                  Mayank Singh},
  title        = {TabLeX: {A} Benchmark Dataset for Structure and Content Information
                  Extraction from Scientific Tables},
  journal      = {CoRR},
  volume       = {abs/2105.06400},
  year         = {2021},
  url          = {https://arxiv.org/abs/2105.06400},
  eprinttype   = {arXiv},
  eprint       = {2105.06400},
  bibsource    = {dblp computer science bibliography, https://dblp.org}
}

@inproceedings{cpapers,
  author    = {Anirudh Sundar and Jin Xu and William Gay and Christopher Richardson and Larry Heck},
  title     = {cPAPERS: A Dataset of Situated and Multimodal Interactive Conversations in Scientific Papers},
  booktitle = {Advances in Neural Information Processing Systems 37 (NeurIPS 2024) Datasets and Benchmarks Track},
  year      = {2024},
  doi       = {10.52202/079017-2119},
  url       = {https://proceedings.neurips.cc/paper_files/paper/2024/hash/7a19a9d527ed544d1272f07b0f8f934e-Abstract-Datasets_and_Benchmarks_Track.html}
}

@misc{sciclaimeval,
      title={SciClaimEval: Cross-modal Claim Verification in Scientific Papers}, 
      author={Xanh Ho and Yun-Ang Wu and Sunisth Kumar and Tian Cheng Xia and Florian Boudin and Andre Greiner-Petter and Akiko Aizawa},
      year={2026},
      eprint={2602.07621},
      archivePrefix={arXiv},
      primaryClass={cs.CL},
      url={https://arxiv.org/abs/2602.07621}, 
}

@inproceedings{teds,
  author    = {Xu Zhong and Elaheh ShafieiBavani and Antonio Jimeno Yepes},
  title     = {Image-Based Table Recognition: Data, Model, and Evaluation},
  booktitle = {Computer Vision -- ECCV 2020},
  year      = {2020},
  url       = {https://arxiv.org/abs/1911.10683}
}

@misc{grits,
      title={GriTS: Grid Table Similarity Metric for Table Structure Recognition},
      author={Brandon Smock and Rohith Pesala and Robin Abraham},
      year={2022},
      eprint={2203.12555},
      archivePrefix={arXiv},
      primaryClass={cs.CV},
      url={https://arxiv.org/abs/2203.12555},
}

@misc{extractbench,
      title={ExtractBench: A Benchmark and Evaluation Methodology for Complex Structured Extraction},
      author={Nick Ferguson and Josh Pennington and Narek Beghian and Aravind Mohan and Douwe Kiela and Sheshansh Agrawal and Thien Hang Nguyen},
      year={2026},
      eprint={2602.12247},
      archivePrefix={arXiv},
      primaryClass={cs.LG},
      url={https://arxiv.org/abs/2602.12247},
}

@misc{meta-harness,
      title={Meta-Harness: End-to-End Optimization of Model Harnesses},
      year={2026},
      eprint={2603.28052},
      archivePrefix={arXiv},
      url={https://arxiv.org/abs/2603.28052},
}

@misc{nl-agent-harnesses,
      title={Natural-Language Agent Harnesses},
      year={2026},
      eprint={2603.25723},
      archivePrefix={arXiv},
      url={https://arxiv.org/abs/2603.25723},
}

@misc{harness-native-software-engineering,
      author={Chaitanya S. and collaborators},
      title={Harness-Native Software Engineering: The Control Plane of Coding Agents},
      year={2026},
      url={https://research.chaitanya.science/papers/harness-native-software-engineering},
}

@misc{anthropic-harness-design,
      author={{Anthropic}},
      title={Harness Design for Long-Running Application Development},
      year={2026},
      month={mar},
      day={24},
      url={https://www.anthropic.com/engineering/harness-design-long-running-apps},
}

@misc{karpathy-autoresearch,
      author={Andrej Karpathy},
      title={autoresearch},
      year={2026},
      howpublished={GitHub repository},
      url={https://github.com/karpathy/autoresearch},
}

@misc{self-refine,
      title={Self-Refine: Iterative Refinement with Self-Feedback}, 
      author={Aman Madaan and Niket Tandon and Prakhar Gupta and Skyler Hallinan and Luyu Gao and Sarah Wiegreffe and Uri Alon and Nouha Dziri and Shrimai Prabhumoye and Yiming Yang and Shashank Gupta and Bodhisattwa Prasad Majumder and Katherine Hermann and Sean Welleck and Amir Yazdanbakhsh and Peter Clark},
      year={2023},
      eprint={2303.17651},
      archivePrefix={arXiv},
      primaryClass={cs.CL},
      url={https://arxiv.org/abs/2303.17651}, 
}

@misc{textgrad,
      title={TextGrad: Automatic "Differentiation" via Text}, 
      author={Mert Yuksekgonul and Federico Bianchi and Joseph Boen and Sheng Liu and Zhi Huang and Carlos Guestrin and James Zou},
      year={2024},
      eprint={2406.07496},
      archivePrefix={arXiv},
      primaryClass={cs.CL},
      url={https://arxiv.org/abs/2406.07496}, 
}

@misc{feedback-descent,
      title={Feedback Descent: Open-Ended Text Optimization via Pairwise Comparison}, 
      author={Yoonho Lee and Joseph Boen and Chelsea Finn},
      year={2025},
      eprint={2511.07919},
      archivePrefix={arXiv},
      primaryClass={cs.LG},
      url={https://arxiv.org/abs/2511.07919}, 
}

@misc{adaevolve,
      title={AdaEvolve: Adaptive LLM Driven Zeroth-Order Optimization}, 
      author={Mert Cemri and Shubham Agrawal and Akshat Gupta and Shu Liu and Audrey Cheng and Qiuyang Mang and Ashwin Naren and Lutfi Eren Erdogan and Koushik Sen and Matei Zaharia and Alex Dimakis and Ion Stoica},
      year={2026},
      eprint={2602.20133},
      archivePrefix={arXiv},
      primaryClass={cs.NE},
      url={https://arxiv.org/abs/2602.20133}, 
}

@misc{alphaevolve,
      title={AlphaEvolve: A coding agent for scientific and algorithmic discovery}, 
      author={Alexander Novikov and Ngân Vũ and Marvin Eisenberger and Emilien Dupont and Po-Sen Huang and Adam Zsolt Wagner and Sergey Shirobokov and Borislav Kozlovskii and Francisco J. R. Ruiz and Abbas Mehrabian and M. Pawan Kumar and Abigail See and Swarat Chaudhuri and George Holland and Alex Davies and Sebastian Nowozin and Pushmeet Kohli and Matej Balog},
      year={2025},
      eprint={2506.13131},
      archivePrefix={arXiv},
      primaryClass={cs.AI},
      url={https://arxiv.org/abs/2506.13131}, 
}

@misc{gepa,
      title={GEPA: Reflective Prompt Evolution Can Outperform Reinforcement Learning}, 
      author={Lakshya A Agrawal and Shangyin Tan and Dilara Soylu and Noah Ziems and Rishi Khare and Krista Opsahl-Ong and Arnav Singhvi and Herumb Shandilya and Michael J Ryan and Meng Jiang and Christopher Potts and Koushik Sen and Alexandros G. Dimakis and Ion Stoica and Dan Klein and Matei Zaharia and Omar Khattab},
      year={2026},
      eprint={2507.19457},
      archivePrefix={arXiv},
      primaryClass={cs.CL},
      url={https://arxiv.org/abs/2507.19457}, 
}

@misc{dspy,
      title={DSPy: Compiling Declarative Language Model Calls into Self-Improving Pipelines}, 
      author={Omar Khattab and Arnav Singhvi and Paridhi Maheshwari and Zhiyuan Zhang and Keshav Santhanam and Sri Vardhamanan and Saiful Haq and Ashutosh Sharma and Thomas T. Joshi and Hanna Moazam and Heather Miller and Matei Zaharia and Christopher Potts},
      year={2023},
      eprint={2310.03714},
      archivePrefix={arXiv},
      primaryClass={cs.CL},
      url={https://arxiv.org/abs/2310.03714}, 
}

@misc{pmc_open_access_subset,
  title        = {PMC Open Access Subset},
  author       = {{National Library of Medicine}},
  year         = {2003},
  note         = {Last modified September 11, 2025},
  howpublished = {\url{https://pmc.ncbi.nlm.nih.gov/tools/openftlist/}},
  institution  = {National Institutes of Health},
}

@misc{terminalbench,
      title={Terminal-Bench: Benchmarking Agents on Hard, Realistic Tasks in Command Line Interfaces}, 
      author={Mike A. Merrill and Alexander G. Shaw and Nicholas Carlini and Boxuan Li and Harsh Raj and Ivan Bercovich and Lin Shi and Jeong Yeon Shin and Thomas Walshe and E. Kelly Buchanan and Junhong Shen and Guanghao Ye and Haowei Lin and Jason Poulos and Maoyu Wang and Marianna Nezhurina and Jenia Jitsev and Di Lu and Orfeas Menis Mastromichalakis and Zhiwei Xu and Zizhao Chen and Yue Liu and Robert Zhang and Leon Liangyu Chen and Anurag Kashyap and Jan-Lucas Uslu and Jeffrey Li and Jianbo Wu and Minghao Yan and Song Bian and Vedang Sharma and Ke Sun and Steven Dillmann and Akshay Anand and Andrew Lanpouthakoun and Bardia Koopah and Changran Hu and Etash Guha and Gabriel H. S. Dreiman and Jiacheng Zhu and Karl Krauth and Li Zhong and Niklas Muennighoff and Robert Amanfu and Shangyin Tan and Shreyas Pimpalgaonkar and Tushar Aggarwal and Xiangning Lin and Xin Lan and Xuandong Zhao and Yiqing Liang and Yuanli Wang and Zilong Wang and Changzhi Zhou and David Heineman and Hange Liu and Harsh Trivedi and John Yang and Junhong Lin and Manish Shetty and Michael Yang and Nabil Omi and Negin Raoof and Shanda Li and Terry Yue Zhuo and Wuwei Lin and Yiwei Dai and Yuxin Wang and Wenhao Chai and Shang Zhou and Dariush Wahdany and Ziyu She and Jiaming Hu and Zhikang Dong and Yuxuan Zhu and Sasha Cui and Ahson Saiyed and Arinbjörn Kolbeinsson and Jesse Hu and Christopher Michael Rytting and Ryan Marten and Yixin Wang and Alex Dimakis and Andy Konwinski and Ludwig Schmidt},
      year={2026},
      eprint={2601.11868},
      archivePrefix={arXiv},
      primaryClass={cs.SE},
      url={https://arxiv.org/abs/2601.11868}, 
}

@misc{certara_q42025,
title = {Certara Reports Fourth Quarter 2025 Financial Results; Provides Full Year 2026 Guidance},
author = {{Certara, Inc.}},
year = {2026},
month = feb,
day = {26},
howpublished = {SEC Exhibit 99.1 earnings release},
url = {https://www.sec.gov/Archives/edgar/data/1827090/000182709026000008/q42025earningsreleaseex99.htm}
}

@inproceedings{bai-etal-2024-schema,
    title = "Schema-Driven Information Extraction from Heterogeneous Tables",
    author = "Bai, Fan  and
      Kang, Junmo  and
      Stanovsky, Gabriel  and
      Freitag, Dayne  and
      Dredze, Mark  and
      Ritter, Alan",
    editor = "Al-Onaizan, Yaser  and
      Bansal, Mohit  and
      Chen, Yun-Nung",
    booktitle = "Findings of the Association for Computational Linguistics: EMNLP 2024",
    month = nov,
    year = "2024",
    address = "Miami, Florida, USA",
    publisher = "Association for Computational Linguistics",
    url = "https://aclanthology.org/2024.findings-emnlp.600/",
    doi = "10.18653/v1/2024.findings-emnlp.600",
    pages = "10252--10273"
}

@inproceedings{padmakumar-etal-2025-intent,
    title = "Intent-aware Schema Generation and Refinement for Literature Review Tables",
    author = "Padmakumar, Vishakh  and
      Chang, Joseph Chee  and
      Lo, Kyle  and
      Downey, Doug  and
      Naik, Aakanksha",
    editor = "Christodoulopoulos, Christos  and
      Chakraborty, Tanmoy  and
      Rose, Carolyn  and
      Peng, Violet",
    booktitle = "Findings of the Association for Computational Linguistics: EMNLP 2025",
    month = nov,
    year = "2025",
    address = "Suzhou, China",
    publisher = "Association for Computational Linguistics",
    url = "https://aclanthology.org/2025.findings-emnlp.1274/",
    doi = "10.18653/v1/2025.findings-emnlp.1274",
    pages = "23450--23472",
    ISBN = "979-8-89176-335-7"
}

@article{chan2022applications,
  title={Applications of Model-Based Meta-Analysis in Drug Development: Chan, Peskov and Song},
  author={Chan, Phyllis and Peskov, Kirill and Song, Xuyang},
  journal={Pharmaceutical Research},
  volume={39},
  number={8},
  pages={1761--1777},
  year={2022},
  publisher={Springer}
}

@article{asai2026synthesizing,
  title={Synthesizing scientific literature with retrieval-augmented language models},
  author={Asai, Akari and He, Jacqueline and Shao, Rulin and Shi, Weijia and Singh, Amanpreet and Chang, Joseph Chee and Lo, Kyle and Soldaini, Luca and Feldman, Sergey and D’Arcy, Mike and others},
  journal={Nature},
  pages={1--7},
  year={2026},
  publisher={Nature Publishing Group UK London}
}

@inproceedings{ghosh-etal-2024-toward,
    title = "Toward Reliable Ad-hoc Scientific Information Extraction: A Case Study on Two Materials Dataset",
    author = "Ghosh, Satanu  and
      Brodnik, Neal  and
      Frey, Carolina  and
      Holgate, Collin  and
      Pollock, Tresa  and
      Daly, Samantha  and
      Carton, Samuel",
    editor = "Ku, Lun-Wei  and
      Martins, Andre  and
      Srikumar, Vivek",
    booktitle = "Findings of the Association for Computational Linguistics: ACL 2024",
    month = aug,
    year = "2024",
    address = "Bangkok, Thailand",
    publisher = "Association for Computational Linguistics",
    url = "https://aclanthology.org/2024.findings-acl.897/",
    doi = "10.18653/v1/2024.findings-acl.897",
    pages = "15109--15123"
}

@article{lu2012biocuration,
  title={Biocuration workflows and text mining: overview of the BioCreative 2012 Workshop Track II},
  author={Lu, Zhiyong and Hirschman, Lynette},
  journal={Database},
  volume={2012},
  pages={bas043},
  year={2012},
  publisher={Oxford University Press}
}

@article{arighi2011biocreative,
  title={BioCreative III interactive task: an overview},
  author={Arighi, Cecilia N and Roberts, Phoebe M and Agarwal, Shashank and Bhattacharya, Sanmitra and Cesareni, Gianni and Chatr-Aryamontri, Andrew and Clematide, Simon and Gaudet, Pascale and Giglio, Michelle Gwinn and Harrow, Ian and others},
  journal={BMC bioinformatics},
  volume={12},
  number={Suppl 8},
  pages={S4},
  year={2011},
  publisher={Springer}
}

@inproceedings{lee2023qasa,
  title={Qasa: advanced question answering on scientific articles},
  author={Lee, Yoonjoo and Lee, Kyungjae and Park, Sunghyun and Hwang, Dasol and Kim, Jaehyeon and Lee, Hong-in and Lee, Moontae},
  booktitle={International Conference on Machine Learning},
  pages={19036--19052},
  year={2023},
  organization={PMLR}
}

@inproceedings{dasigi-etal-2021-dataset,
    title = "A Dataset of Information-Seeking Questions and Answers Anchored in Research Papers",
    author = "Dasigi, Pradeep  and
      Lo, Kyle  and
      Beltagy, Iz  and
      Cohan, Arman  and
      Smith, Noah A.  and
      Gardner, Matt",
    editor = "Toutanova, Kristina  and
      Rumshisky, Anna  and
      Zettlemoyer, Luke  and
      Hakkani-Tur, Dilek  and
      Beltagy, Iz  and
      Bethard, Steven  and
      Cotterell, Ryan  and
      Chakraborty, Tanmoy  and
      Zhou, Yichao",
    booktitle = "Proceedings of the 2021 Conference of the North American Chapter of the Association for Computational Linguistics: Human Language Technologies",
    month = jun,
    year = "2021",
    address = "Online",
    publisher = "Association for Computational Linguistics",
    url = "https://aclanthology.org/2021.naacl-main.365/",
    doi = "10.18653/v1/2021.naacl-main.365",
    pages = "4599--4610"
}

@inproceedings{zhao-etal-2025-multimodal-foundation,
    title = "Can Multimodal Foundation Models Understand Schematic Diagrams? An Empirical Study on Information-Seeking {QA} over Scientific Papers",
    author = "Zhao, Yilun  and
      Wang, Chengye  and
      Li, Chuhan  and
      Cohan, Arman",
    editor = "Che, Wanxiang  and
      Nabende, Joyce  and
      Shutova, Ekaterina  and
      Pilehvar, Mohammad Taher",
    booktitle = "Findings of the Association for Computational Linguistics: ACL 2025",
    month = jul,
    year = "2025",
    address = "Vienna, Austria",
    publisher = "Association for Computational Linguistics",
    url = "https://aclanthology.org/2025.findings-acl.957/",
    doi = "10.18653/v1/2025.findings-acl.957",
    pages = "18598--18631",
    ISBN = "979-8-89176-256-5"
}

@inproceedings{wadden-etal-2020-fact,
    title = "Fact or Fiction: Verifying Scientific Claims",
    author = "Wadden, David  and
      Lin, Shanchuan  and
      Lo, Kyle  and
      Wang, Lucy Lu  and
      van Zuylen, Madeleine  and
      Cohan, Arman  and
      Hajishirzi, Hannaneh",
    editor = "Webber, Bonnie  and
      Cohn, Trevor  and
      He, Yulan  and
      Liu, Yang",
    booktitle = "Proceedings of the 2020 Conference on Empirical Methods in Natural Language Processing (EMNLP)",
    month = nov,
    year = "2020",
    address = "Online",
    publisher = "Association for Computational Linguistics",
    url = "https://aclanthology.org/2020.emnlp-main.609/",
    doi = "10.18653/v1/2020.emnlp-main.609",
    pages = "7534--7550"
}

@article{wang2025foundation,
  title={A foundation model for human-AI collaboration in medical literature mining},
  author={Wang, Zifeng and Cao, Lang and Jin, Qiao and Chan, Joey and Wan, Nicholas and Afzali, Behdad and Cho, Hyun-Jin and Choi, Chang-In and Emamverdi, Mehdi and Gill, Manjot K and others},
  journal={Nature communications},
  volume={16},
  number={1},
  pages={8361},
  year={2025},
  publisher={Nature Publishing Group UK London}
}

@article{liu2025metascale,
  title={Metascale: Test-time scaling with evolving meta-thoughts},
  author={Liu, Qin and Zhou, Wenxuan and Xu, Nan and Huang, James Y and Wang, Fei and Zhang, Sheng and Poon, Hoifung and Chen, Muhao},
  journal={arXiv preprint arXiv:2503.13447},
  year={2025}
}

@inproceedings{zhang-etal-2024-scier,
    title = "{S}ci{ER}: An Entity and Relation Extraction Dataset for Datasets, Methods, and Tasks in Scientific Documents",
    author = "Zhang, Qi  and
      Chen, Zhijia  and
      Pan, Huitong  and
      Caragea, Cornelia  and
      Latecki, Longin Jan  and
      Dragut, Eduard",
    editor = "Al-Onaizan, Yaser  and
      Bansal, Mohit  and
      Chen, Yun-Nung",
    booktitle = "Proceedings of the 2024 Conference on Empirical Methods in Natural Language Processing",
    month = nov,
    year = "2024",
    address = "Miami, Florida, USA",
    publisher = "Association for Computational Linguistics",
    url = "https://aclanthology.org/2024.emnlp-main.726/",
    doi = "10.18653/v1/2024.emnlp-main.726",
    pages = "13083--13100"
}

@inproceedings{duan-etal-2025-scinlp,
    title = "{S}ci{NLP}: A Domain-Specific Benchmark for Full-Text Scientific Entity and Relation Extraction in {NLP}",
    author = "Duan, Decheng  and
      Peng, Jitong  and
      Zhang, Yingyi  and
      Zhang, Chengzhi",
    editor = "Christodoulopoulos, Christos  and
      Chakraborty, Tanmoy  and
      Rose, Carolyn  and
      Peng, Violet",
    booktitle = "Proceedings of the 2025 Conference on Empirical Methods in Natural Language Processing",
    month = nov,
    year = "2025",
    address = "Suzhou, China",
    publisher = "Association for Computational Linguistics",
    url = "https://aclanthology.org/2025.emnlp-main.732/",
    doi = "10.18653/v1/2025.emnlp-main.732",
    pages = "14473--14486",
    ISBN = "979-8-89176-332-6"
}

@inproceedings{jain-etal-2020-scirex,
    title = "{S}ci{REX}: {A} Challenge Dataset for Document-Level Information Extraction",
    author = "Jain, Sarthak  and
      van Zuylen, Madeleine  and
      Hajishirzi, Hannaneh  and
      Beltagy, Iz",
    editor = "Jurafsky, Dan  and
      Chai, Joyce  and
      Schluter, Natalie  and
      Tetreault, Joel",
    booktitle = "Proceedings of the 58th Annual Meeting of the Association for Computational Linguistics",
    month = jul,
    year = "2020",
    address = "Online",
    publisher = "Association for Computational Linguistics",
    url = "https://aclanthology.org/2020.acl-main.670/",
    doi = "10.18653/v1/2020.acl-main.670",
    pages = "7506--7516"
}

@inproceedings{luan-etal-2018-multi,
    title = "Multi-Task Identification of Entities, Relations, and Coreference for Scientific Knowledge Graph Construction",
    author = "Luan, Yi  and
      He, Luheng  and
      Ostendorf, Mari  and
      Hajishirzi, Hannaneh",
    editor = "Riloff, Ellen  and
      Chiang, David  and
      Hockenmaier, Julia  and
      Tsujii, Jun{'}ichi",
    booktitle = "Proceedings of the 2018 Conference on Empirical Methods in Natural Language Processing",
    month = oct # "-" # nov,
    year = "2018",
    address = "Brussels, Belgium",
    publisher = "Association for Computational Linguistics",
    url = "https://aclanthology.org/D18-1360/",
    doi = "10.18653/v1/D18-1360",
    pages = "3219--3232"
}

@article{luo2022biored,
  title={BioRED: a rich biomedical relation extraction dataset},
  author={Luo, Ling and Lai, Po-Ting and Wei, Chih-Hsuan and Arighi, Cecilia N and Lu, Zhiyong},
  journal={Briefings in Bioinformatics},
  volume={23},
  number={5},
  pages={bbac282},
  year={2022},
  publisher={Oxford University Press}
}

@inproceedings{nye-etal-2018-corpus,
    title = "A Corpus with Multi-Level Annotations of Patients, Interventions and Outcomes to Support Language Processing for Medical Literature",
    author = "Nye, Benjamin  and
      Li, Junyi Jessy  and
      Patel, Roma  and
      Yang, Yinfei  and
      Marshall, Iain  and
      Nenkova, Ani  and
      Wallace, Byron",
    editor = "Gurevych, Iryna  and
      Miyao, Yusuke",
    booktitle = "Proceedings of the 56th Annual Meeting of the Association for Computational Linguistics (Volume 1: Long Papers)",
    month = jul,
    year = "2018",
    address = "Melbourne, Australia",
    publisher = "Association for Computational Linguistics",
    url = "https://aclanthology.org/P18-1019/",
    doi = "10.18653/v1/P18-1019",
    pages = "197--207"
}


\newpage
\appendix

\section{Appendix}
\label{sec:appendix}

\begin{figure*}[!ht]
    \centering
    \includegraphics[width=\textwidth]{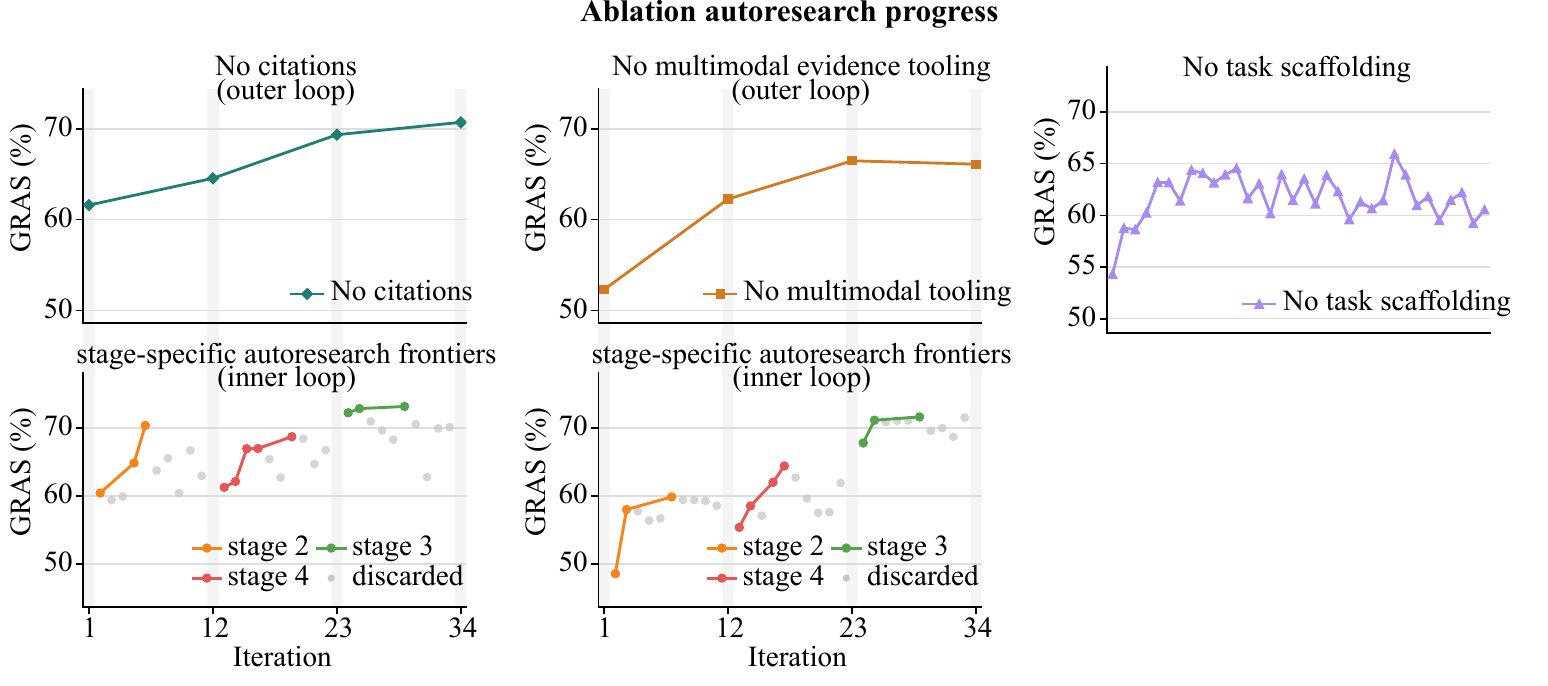}
    \caption{Autoresearch trajectories for ablated harnesses. Top panels show outer-loop progress over the autoresearch budget, while bottom panels show stage-specific inner-loop candidates, with retained frontier updates separated from discarded attempts. Without task scaffolding, the method lacks stable stage-localized workspaces, so only the aggregate progress trajectory is shown.}
    \label{fig:ablation-progress-2}
\end{figure*}

\paragraph{Ablation Autoresearch Trajectories.}
\Cref{fig:ablation-progress-2} shows the optimization trajectories for each ablated harness. For the no-provenance-traces and no-multimodal-tooling variants, the top panels show outer-loop improvements over the autoresearch budget, while the bottom panels show stage-specific inner-loop candidates, distinguishing retained frontiers from discarded attempts. The no-task-scaffolding variant lacks stage-localized frontiers because the harness no longer decomposes execution into stable stages. These trajectories illustrate that autoresearch can still find improvements under ablations, but the search becomes less effective or less structured when key harness components are removed.

\subsection{Representative Harness Changes}
\label{app:representative-revisions}

This subsection provides three representative harness changes discovered by \ours.
Taken together, these changes clarify why autoresearch helps in this setting: the gains come from improving how the harness preserves scientific structure, accesses multimodal evidence, and normalizes extracted values.

\paragraph{Figure-grounded evidence access.}
A representative change to the figure-reading harness replaced general visual inspection with explicit axis calibration and narrow-band marker detection. The resulting workflow first anchors the plot axes in pixel space, converts between pixel rows and numeric values, and only then reads candidate points near the relevant timepoint. The mechanism is illustrated by the following representative snippet:

\begin{verbatim}
# gray: single-channel image intensity array used for scanning
# x_ticks: x-pixel centers for labeled visits or timepoints
# THRESHOLD: darkness cutoff for axis, tick, and marker evidence
# AXIS_* and TICK_*: approximate search windows around the plot axes
# *_DARK_MIN: minimum dark-pixel evidence for accepting a local candidate

AXIS_COL_MIN = plot_left_col
AXIS_COL_MAX = plot_left_col + axis_search_width
AXIS_SCAN_ROW_MIN, AXIS_SCAN_ROW_MAX = plot_top_row, plot_bottom_row
TICK_ROW_MIN, TICK_ROW_MAX = plot_top_row, plot_bottom_row
TICK_BAND_WIDTH = axis_tick_band_width
MARKER_BAND_RADIUS = marker_scan_radius
AXIS_DARK_MIN = axis_dark_min
TICK_DARK_MIN = tick_dark_min
MARKER_DARK_MIN = marker_dark_min

# Search candidate columns for the y-axis line.
axis_col = None
for c in range(AXIS_COL_MIN, AXIS_COL_MAX):
    dark = np.sum(gray[AXIS_SCAN_ROW_MIN:AXIS_SCAN_ROW_MAX, c] < THRESHOLD)
    if dark > AXIS_DARK_MIN:
        axis_col = c

# Search just left of the detected axis for y-tick candidates.
tick_rows = []
for r in range(TICK_ROW_MIN, TICK_ROW_MAX):
    dark = np.sum(gray[r, axis_col-TICK_BAND_WIDTH:axis_col] < THRESHOLD)
    if dark >= TICK_DARK_MIN:
        tick_rows.append(r)

pixels_per_unit = (px_bottom_tick - px_top_tick) \
    / (val_top_tick - val_bottom_tick)

def ypx_to_val(y_px):
    return (y_zero_px - y_px) / pixels_per_unit

# Scan a narrow band around each x-position for candidate data markers.
for label, xcol in x_ticks.items():
    x_lo = max(0, xcol - MARKER_BAND_RADIUS)
    x_hi = min(w, xcol + MARKER_BAND_RADIUS + 1)
    candidate_rows = []
    for r in range(plot_top_row, plot_bottom_row):
        dark_count = np.sum(gray[r, x_lo:x_hi] < THRESHOLD)
        if dark_count >= MARKER_DARK_MIN:
            candidate_rows.append((r, dark_count))

    groups = group_consecutive(candidate_rows, max_gap=row_group_gap)
    for g in groups:
        center_row = best_row_in_group(g)
        y_val = ypx_to_val(center_row)
        print(f"{label}: row {center_row}, y = {y_val:.2f}")
\end{verbatim}

This change matters because figure-driven rows are often lost even when the correct figure has been found: the real bottleneck is turning a visually localized trace into a stable numeric readout aligned to the correct visit and arm.

\paragraph{Family preservation.}
A second class of changes are applied to how \texttt{baseline} related attributes are emitted before endpoint-level expansion.
This family-preservation logic reduces a recurring class of failures where pooled rows, subgroup rows, or comparison families are merged too early and cannot be recovered downstream.
Due to length considerations, we present the pseudocode below instead of a full code listing:

\begin{algorithm}[!ht]
\caption{Stage-$n$ Row-Family Preservation}
\label{alg:stage3-row-family-preservation}
\begin{algorithmic}[1]
\Require paper $p$, carried-forward row $\hat{r}^{(n-1)} \in \hat{Y}^{(n-1)}$, evidence artifacts $(\mathcal{T}(p), \mathcal{B}(p), \mathcal{G}(p), \mathcal{A}(p))$
\Ensure emitted row set $\mathcal{R}^{(n)}(\hat{r}^{(n-1)}) \subseteq \hat{Y}^{(n)}$ over $\mathcal{F}^{(n)}$
\State $\mathcal{R}^{(n)}(\hat{r}^{(n-1)}) \gets \emptyset$
\State $\mathcal{B}_{\mathrm{base}}(p, \hat{r}^{(n-1)}) \gets$ baseline-characteristics table candidates for the endpoint in $\hat{r}^{(n-1)}$
\State $\mathcal{B}_{\mathrm{comp}}(p, \hat{r}^{(n-1)}) \gets$ later comparison-table candidates for the endpoint in $\hat{r}^{(n-1)}$
\State $\mathcal{B}^\star(p, \hat{r}^{(n-1)}) \gets \mathcal{B}_{\mathrm{base}}(p, \hat{r}^{(n-1)})$ if both candidate sets are non-empty; otherwise use the available candidate set
\State $\Pi(\hat{r}^{(n-1)}, p) \gets$ explicit analysis families supported by the paper for $\hat{r}^{(n-1)}$
\If{$|\Pi(\hat{r}^{(n-1)}, p)| = 0$}
    \State $\Pi(\hat{r}^{(n-1)}, p) \gets \{f_0\}$ \Comment{default carried-forward family}
\EndIf
\ForAll{$f \in \Pi(\hat{r}^{(n-1)}, p)$}
    \State initialize $\hat{r}^{(n)}_f$ from the carried-forward attributes in $\hat{r}^{(n-1)}$
    \State populate \texttt{baseline}$(\hat{r}^{(n)}_f)$ and \texttt{baseline.sd}$(\hat{r}^{(n)}_f)$ from $\mathcal{B}^\star(p, \hat{r}^{(n-1)})$ for family $f$
    \If{\texttt{n.randomized}$(\hat{r}^{(n-1)})$ is blank $\land$ no exact-family denominator is reported in $(\mathcal{T}(p), \mathcal{B}(p), \mathcal{G}(p), \mathcal{A}(p))$}
        \State preserve blank \texttt{n.randomized}$(\hat{r}^{(n)}_f)$
    \Else
        \State set \texttt{n.randomized}$(\hat{r}^{(n)}_f)$ from the exact-family denominator when available
    \EndIf
    \State carry the canonical endpoint spelling and family identity from $\hat{r}^{(n-1)}$ into $\hat{r}^{(n)}_f$
    \State $\mathcal{R}^{(n)}(\hat{r}^{(n-1)}) \gets \mathcal{R}^{(n)}(\hat{r}^{(n-1)}) \cup \{\hat{r}^{(n)}_f\}$
\EndFor
\State \Return $\mathcal{R}^{(n)}(\hat{r}^{(n-1)})$
\end{algorithmic}
\end{algorithm}

\paragraph{Deterministic postprocessing.}
A third class of changes adds deterministic same-row cleanup after the raw extraction. Rather than relying on the model to emit every derived value directly, the harness now fills outcome attributes such as \texttt{value}, \texttt{percentchange}, and \texttt{n.observed} when the same row already contains enough local evidence to do so consistently. The following excerpt from the postprocessor shows the core mechanism:

\begin{verbatim}
if baseline is not None and change is not None:
    if _is_blank(row.get("value")):
        row["value"] = _format_decimal(baseline + change)
    if baseline != 0 and _is_blank(row.get("percentchange")):
        row["percentchange"] = _format_decimal(
            (change / baseline) * Decimal("100")
        )

if n_randomized is not None and _is_blank(row.get("n.observed")):
    row["n.observed"] = _format_decimal(n_randomized)
\end{verbatim}

\subsection{Failure mode case studies}
\label{app:representative-failure-cases}

The case studies below elaborate the three contrasting case types summarized in \cref{sec:analysis}, with per-paper GRAS scores averaged across the three replicate runs of each family. \Cref{tab:appendix-failure-cases} collects the headline numbers for the four representative papers discussed in this section.

\begin{table}[!ht]
\centering
\small
\setlength{\tabcolsep}{4pt}
\renewcommand{\arraystretch}{1.08}
\caption{Per-paper GRAS for the four papers used as failure-mode case studies, averaged across three replicate runs per family. The ``Baseline Agent (best)'' column reports the strongest baseline agent; ``Baseline Agents (avg)'' averages all baseline agents.}
\label{tab:appendix-failure-cases}
\begin{tabular}{lcccccc}
\toprule
\textbf{PMID} & \textbf{\ours} & \textbf{Baseline Agent} & \textbf{Baseline Agents} & \textbf{No task} & \textbf{No} & \textbf{No multimodal} \\
& & (best) & (avg) & \textbf{scaffolding} &  \textbf{provenances} & \textbf{evidence tooling} \\
\midrule
27756421 & 0.821 & 0.302 (Codex) & 0.271 & 0.310 & 0.580 & 0.503 \\
22291741 & 1.000 & 0.875 (Copilot) & 0.716 & 0.460 & 0.714 & 0.714 \\
31884472 & 0.755 & 0.720 (Codex) & 0.636 & 0.781 & 0.356 & 0.364 \\
29154277 & 0.478 & 0.317 (Codex) & 0.282 & 0.403 & 0.543 & 0.484 \\
\bottomrule
\end{tabular}
\end{table}

\begin{figure}[!ht]
\centering
\begin{minipage}[c]{0.48\textwidth}
\centering
\includegraphics[width=\linewidth]{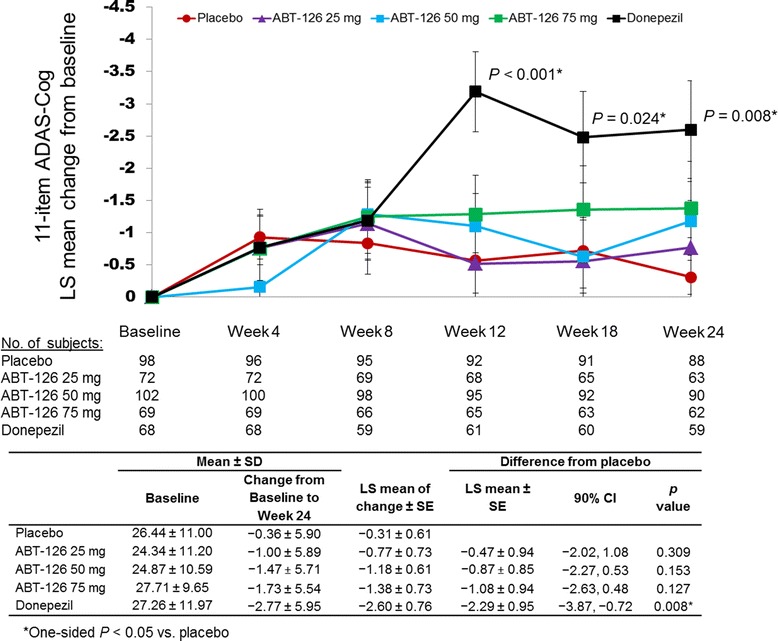}
\\\small (a) Original Fig.~2 in the paper.
\end{minipage}\hfill
\begin{minipage}[c]{0.48\textwidth}
\centering
\includegraphics[width=\linewidth]{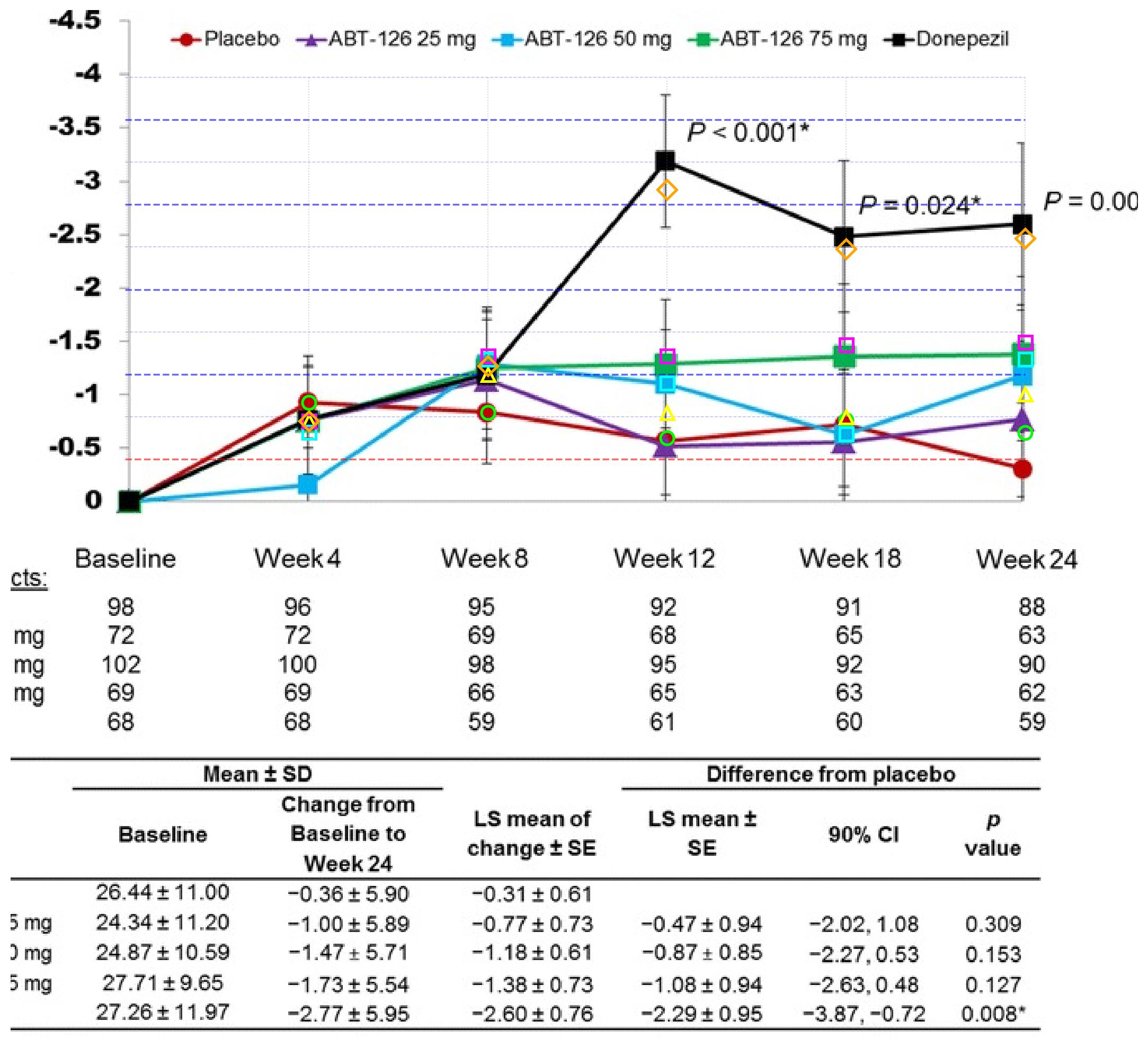}
\\\small (b) The same figure after \ours's multimodal evidence tooling.
\end{minipage}
\caption{Case A (\href{https://pubmed.ncbi.nlm.nih.gov/27756421/}{PMID 27756421}). Panel (a) is the original ADAS-Cog change-from-baseline trajectory plotted across visits per treatment arm; panel (b) is the artifact written by \ours's multimodal evidence tooling, with calibrated y-axis gridlines overlaid on the marker band and the per-visit (mean, SE, $n$) read-out re-emitted as a structured table beneath the plot. Off-the-shelf baselines and the no-multimodal ablation either skip the figure or estimate values without calibration; \ours uses the gridline overlay to produce the 80 per-visit rows that align with gold.}
\label{fig:case-a-27756421}
\end{figure}

\paragraph{Case A: Baselines fail, \ours succeeds (\href{https://pubmed.ncbi.nlm.nih.gov/27756421/}{PMID 27756421}).}
This trial reports its primary ADAS-Cog outcomes as a multi-visit figure (\Cref{fig:case-a-27756421} left), with per-visit means as plotted markers and per-visit subject counts in a row beneath the plot. The gold reference contains 80 rows spanning all visits, treatment arms, and endpoint variants. Every baseline agent reports only the final-visit row per arm: codex emits 15 rows, meta-harness 10, and copilot variants between 10 and 30, all clustered around GRAS~\(\approx 0.27\). \ours emits 80 rows aligned to the gold reference and reaches \(0.821\), a \(+0.52\) gap over the strongest baseline. The supporting provenances explicitly reference axis-calibrated gridline overlays produced by the multimodal evidence tooling (e.g., a per-row note ``LS mean change estimated from calibrated gridline reading of the plotted marker''); \Cref{fig:case-a-27756421} shows the original figure beside the version produced by \ours using multimodal evidence tooling.

\paragraph{Case B: Ablations fail, \ours succeeds.}
We pick one PMID per ablation family that exposes the load-bearing role of the corresponding \ours component.

\begin{figure}[!ht]
\centering
\includegraphics[width=0.85\textwidth]{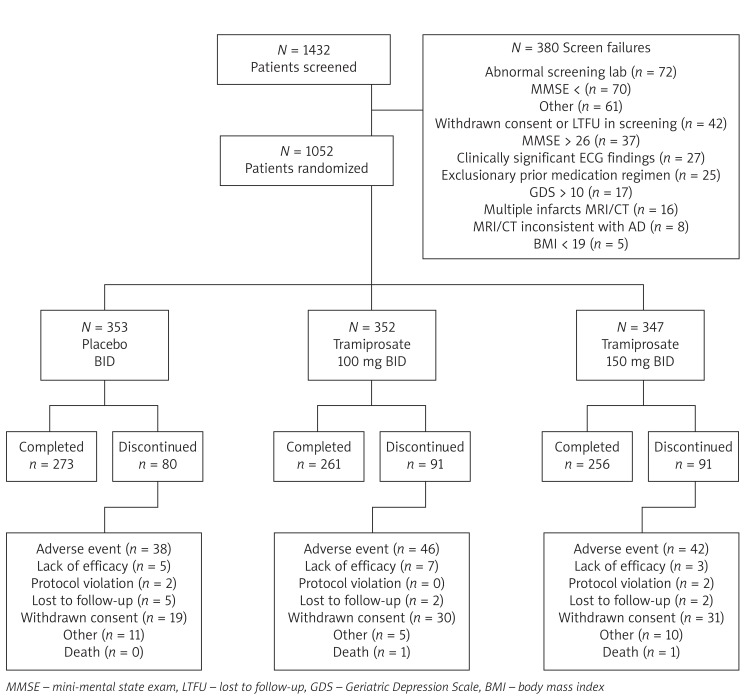}
\caption{Case B / no-task-scaffolding (\href{https://pubmed.ncbi.nlm.nih.gov/22291741/}{PMID 22291741}). The paper's CONSORT diagram fixes the trial structure: 1052 patients randomized to placebo and two tramiprosate doses (\(n = 353, 352, 347\)), with MMSE at week~78 as the gold-relevant primary endpoint. \ours emits the three matching rows; the no-scaffolding ablation instead enumerates 72 visit\,$\times$\,endpoint rows from the ADAS-Cog tables because no prior-stage carry-forward keeps it on the MMSE-week-78 family.}
\label{fig:case-b-22291741}
\end{figure}

\textit{No-task-scaffolding (\href{https://pubmed.ncbi.nlm.nih.gov/22291741/}{PMID 22291741}).}~The trial's primary endpoint is MMSE at week~78 across three arms (placebo and two tramiprosate doses), as fixed by the CONSORT diagram in \Cref{fig:case-b-22291741}; the gold reference is just three rows. \ours emits exactly those three rows and matches gold on every scored field (\(1.000\)). Removing task scaffolding causes the agent to drop the staged endpoint-family carry-forward and instead enumerate every visit \(\times\) endpoint pair from the body's ADAS-Cog tables (72 rows), producing a \(0.460\) score.

\textit{No-provenances (\href{https://pubmed.ncbi.nlm.nih.gov/31884472/}{PMID 31884472}).}~The gold layout has 11 rows spread across four dose levels: per dose, one primary ``ADAS-cog-11~$\vert$~MMSE'' row plus one or two ``ADAS-cog-11''-only subgroup rows distinguished by which subset of randomized subjects had each rating. \ours preserves this row family (12 rows; \(0.755\)) by anchoring each subgroup row in a distinct provenance evidence span (\texttt{s2c001} through \texttt{s2c021}). Stripping provenances collapses the output to four primary rows (\(0.356\)). The same paper is the strongest no-multimodal regression as well (\(0.364\)) because the subgroup decomposition is communicated in a small inset table whose structural reading benefits from multimodal tooling.

\textit{No-multimodal (PMID 27756421, stage 4).}~Re-using the trial from Case A, removing multimodal tooling does not eliminate figure reading entirely (the ablation still emits 50 rows), but it replaces the calibrated gridline reads shown in \Cref{fig:case-a-27756421}(b) with prose ``visually estimated from the plotted marker'' and loses the per-visit subject counts. GRAS drops from \(0.821\) to \(0.503\), confirming that the gain from multimodal tooling is predominantly numerical fidelity rather than figure detection.

\paragraph{Case C: \ours still fails (\href{https://pubmed.ncbi.nlm.nih.gov/29154277/}{PMID 29154277}).}
The trial reports baseline cognitive scores under multiple analysis populations within a single subgroup: the gold reference contains 31 rows across two doses, two endpoints (ADAS-cog-11 and MMSE), and four subgroups, with each (subgroup, endpoint) cell reporting one to three baseline values from different analysis populations (e.g., FAS, observed-cases, completers). \ours emits only 14 rows (one per (dose, subgroup, endpoint)), collapsing the analysis-population variants. GRAS is consistently \(0.48 \pm 0.02\) across replicates. Closing this gap requires the harness to recognise the analysis-population axis as a fourth row-family dimension, which the current pipeline does not.

Taken together, these cases show that \ours's headline gains over baselines come predominantly from figure trajectory reconstruction (Case A) and stagewise row-family preservation (Case B), and that the largest remaining gap is along an axis (analysis-population) that the current harness does not yet treat as a row-distinguishing field (Case C).

\subsection{Evaluation Papers}
\label{app:evaluation-papers}

Table~\ref{tab:appendix-evaluation-papers} reports the 23 PMIDs in the evaluation set and marks the 8 papers used in the development subset inline.

\begin{table}[!ht]
\centering
\small
\setlength{\tabcolsep}{6pt}
\renewcommand{\arraystretch}{1.08}
\newcommand{\appcmark}{\textcolor{green!60!black}{\ensuremath{\checkmark}}}
\begin{tabular}{p{0.14\textwidth}p{0.76\textwidth}}
\toprule
\textbf{PMIDs} & 29154277 (\appcmark), 29067345 (\appcmark), 26064192 (\appcmark), 31329216, 32508323, 27176461, 28550255, 27756421 (\appcmark), 22291741 (\appcmark), 29067304 (\appcmark), 31884472 (\appcmark), 30231896, 30134967, 22606044, 32568196, 31944580, 30309389, 29221491, 29695589, 18213383, 25755685 (\appcmark), 22567095, 24423155 \\
\bottomrule
\end{tabular}
\caption{PMIDs for the 23-paper evaluation set. A checkmark in parentheses marks membership in the 8-paper development subset.}
\label{tab:appendix-evaluation-papers}
\end{table}

\subsection{Schema Details}
\label{app:schema-details}

Table~\ref{tab:appendix-scored-fields} gives the 20 scored attributes used in the main experiments.

\begin{table*}[t]
\centering
\small
\setlength{\tabcolsep}{4pt}
\renewcommand{\arraystretch}{1.08}
\begin{tabular}{p{0.12\textwidth}p{0.23\textwidth}p{0.16\textwidth}p{0.4\textwidth}}
\toprule
\textbf{Category} & \textbf{Attribute} & \textbf{Type} & \textbf{Description} \\
\midrule
Source & \texttt{source.number} & numerical-discrete & Unique ID number of the reference. Use the PMID if available; for other references, assign a unique number. \\
Study & \texttt{registry.study.id} & text-descriptive & Registry ID of the study, including the registry acronym; include multiple IDs when applicable. \\
Study & \texttt{indication} & text-categorical & Indication of the study. \\
Study & \texttt{treatment.duration} & numerical-continuous & Duration of treatment in weeks, corresponding to the end of regular dosing. \\
Study & \texttt{study.duration} & numerical-continuous & Total study duration in weeks, including follow-up. \\
Endpoint & \texttt{endpoint} & text-categorical & Name of the endpoint. \\
Arm & \texttt{n.randomized} & numerical-discrete & Number of patients randomized to the arm or stratum. \\
Treatment & \texttt{drug1} & text-categorical & Name of the first drug; for combinations, the experimental drug is listed first. \\
Treatment & \texttt{drug1.dose} & numerical-continuous & Dose of the first drug, standardized to a common unit across studies. \\
Treatment & \texttt{drug1.dose.unit} & text-descriptive & Standardized dose unit used for the first drug. \\
Endpoint & \texttt{endpoint.unit} & text-categorical & Unit used for baseline, value, and change after endpoint normalization. \\
Endpoint & \texttt{baseline} & numerical-continuous & Baseline value. \\
Endpoint & \texttt{baseline.sd} & numerical-continuous & Standard deviation for the baseline value. \\
Endpoint & \texttt{n.observed} & numerical-discrete & Number of patients analyzed for the endpoint value. \\
Endpoint & \texttt{value} & numerical-continuous & Mean or median endpoint value at the reported timepoint, normalized to the preferred unit. \\
Endpoint & \texttt{value.sd} & numerical-continuous & Standard deviation for the endpoint value. \\
Endpoint & \texttt{change} & numerical-continuous & Mean or median change from baseline at the reported timepoint. \\
Endpoint & \texttt{change.sd} & numerical-continuous & Standard deviation for change from baseline. \\
Endpoint & \texttt{percentchange} & numerical-continuous & Mean or median percent change from baseline at the reported timepoint. \\
Endpoint & \texttt{percentchange.sd} & numerical-continuous & Standard deviation for percent change from baseline. \\
\bottomrule
\end{tabular}
\caption{The 20 scored attributes used in the main experiments.}
\label{tab:appendix-scored-fields}
\end{table*}



\end{document}